\documentclass[review]{elsarticle}

\usepackage{lineno}
\usepackage{amsmath}
\usepackage{amsfonts}
\usepackage{amssymb}
\usepackage{amsthm}
\usepackage{algorithm2e}
\usepackage{multirow}
\usepackage{rotating}
\usepackage{makecell}
\usepackage{subfigure}
\usepackage{color}
\usepackage[top=2.5cm,bottom=3cm,left=3cm,right=2.5cm]{geometry}

\modulolinenumbers[1]
\newtheorem{mydef}{Definition}

\journal{Applied Soft Computing, Volume 52 Pages 909 to 924 }

\begin{document}

\begin{frontmatter}

\title{Ensemble of Heterogeneous Flexible Neural Trees Using Multiobjective Genetic Programming}

\author[address1]{Varun Kumar Ojha\corref{correspondingauthor}}
\cortext[correspondingauthor]{Varun Kumar Ojha}
\ead{varun.kumar.ojha@vsb.cz}
\author[address2]{Ajith Abraham}
\ead{ajith.abraham@ieee.org}

\author[address1]{V{\'{a}}clav Sn{\'{a}}{\v{s}}el}
\ead{vaclav.snasel@vsb.cz}
\address[address1]{IT4Innovations, V{\v{S}}B-Technical University of Ostrava, Ostrava, Czech Republic}
\address[address2]{Machine Intelligence Research Labs (MIR Labs), Auburn, WA, USA}

\begin{abstract}
	Machine learning algorithms are inherently multiobjective in nature, where approximation error minimization and model's complexity simplification are two conflicting objectives. We proposed a multiobjective genetic programming (MOGP) for creating a heterogeneous flexible neural tree (HFNT), tree-like flexible feedforward neural network model. The functional heterogeneity in neural tree nodes was introduced to capture a better insight of data during learning because each input in a dataset possess different features. MOGP guided an initial HFNT population towards Pareto-optimal solutions, where the final population was used for making an ensemble system. A \emph{diversity index} measure along with \emph{approximation error} and \emph{complexity} was introduced to maintain diversity among the candidates in the population. Hence, the ensemble was created by using accurate, structurally simple, and diverse candidates from MOGP final population. Differential evolution algorithm was applied to fine-tune the underlying parameters of the selected candidates. A comprehensive test over classification, regression, and time-series datasets proved the efficiency of the proposed algorithm over other available prediction methods. Moreover, the heterogeneous creation of HFNT proved to be efficient in making ensemble system from the final population.    
\end{abstract}


\end{frontmatter}


\section{Introduction}
\label{sec-intro}
Structure optimization of a feedforward neural network (FNN) and its impact on FNN's generalization ability inspired the flexible neural tree (FNT)~\cite{chen2005time}. FNN components such as weights, structure, and activation function are the potential candidates for the optimization, which improves FNN's generalization ability to a great extent~\cite{yao1997new}. These efforts are notable because of FNN's ability to solve a large range of real-world problems~\cite{basheer2000artificial,maren2014handbook,sethi2014artificial,tkavc2016artificial}. Followings are the significance structure optimization methods: constructive and pruning algorithms~\cite{fahlman1989cascade,nadal1989study}, EPNet~\cite{yao1997new}, NeuroEvolution of Augmenting Topologies~\cite{stanley2002evolving}, sparse neural trees~\cite{zhang1997evolutionary}, Cooperative co-evolution approach~\cite{potter2000cooperative}, etc. Similarly, many efforts focus on the optimization of hybrid training of FNN such as \cite{yaghini2013hybrid,wang2015feed,wang2015fruit}. FNT was an additional step into this series of efforts, which was proposed to evolve as a tree-like feed-forward neural network model, where the probabilistic incremental program evolution (PIPE)~\cite{salustowicz1997probabilistic} was applied optimize the tree structure~\cite{chen2005time}. The underlying parameter vector of the developed FNT (weights associated with the edges and arguments of the activation functions) was optimized by metaheuristic algorithms, which are nature-inspired parameter optimization algorithms~\cite{kar2016bio}. The evolutionary process allowed FNT to select significant input features from an input feature set. 


In the design of FNT, the non-leaf nodes are the computational node, which takes an activation function. Hence, rather than relying on a fixed activation function, if the selection of activation function at the computational nodes is allowed to be selected by the evolutionary process. Then, it produces heterogeneous FNTs (HFNT) with the heterogeneity in its structure, computational nodes, and input set. In addition, heterogeneous function allowed HFNT to capture different feature of the datasets efficiently since each input in the datasets posses different features. The evolutionary process provides adaptation in structure, weights, activation functions, and input features. Therefore, an optimum HFNT is the one that offers the lowest approximation error with the simplest tree structure and the smallest input feature set. However, approximation error minimization and structure simplification are two conflicting objectives~\cite{jin2008pareto}. Hence, a multiobjective evolutionary approach~\cite{deb2001multi} may offer an optimal solution(s) by maintaining a balance between these objectives.       

Moreover, in the proposed work, an evolutionary process guides a population of HFNTs towards Pareto-optimum solutions. Hence, the final population may contain several solutions that are close to the best solution. Therefore, an ensemble system was constructed by exploiting many candidates of the population (candidate, solution, and model are synonymous in this article). Such ensemble system takes advantage of many solutions including the best solution~\cite{yao1998making}. Diversity among the chosen candidates holds the key in making a good ensemble system~\cite{kuncheva2003measures}. Therefore, the solutions in a final population should fulfill the following objectives: low approximation error, structural simplicity, and high diversity. However, these objectives are conflicting to each other. A fast elitist nondominated sorting genetic algorithm (NSGA-II)-based multiobjective genetic programming (MOGP) was employed to guide a population of HFNTs~\cite{deb2000fast}. The underlying parameters of selected models were further optimized by using differential evaluation (DE) algorithm~\cite{das2016recent}. Therefore, we may summarize the key contributions of this work are as follows:
\begin{enumerate}[1)]
	\item A heterogeneous flexible neural tree (HFNT) for function approximation and feature selection was proposed.
	\item HFNT was studied under an NSGA-II-based multiobjective genetic programming framework. Thus, it was termed HFNT$^\text{M}$.
	\item Alongside \emph{approximation error} and \emph{tree size} (complexity), a \emph{diversity index} was introduced to maintain diversity among the candidates in the population.
	\item HFNT$^\text{M}$ was found competitive with other algorithms when compared and cross-validated over classification, regression, and time-series datasets.
	\item The proposed evolutionary weighted ensemble of HFNTs final population further improved its performance.
\end{enumerate}

A detailed literature review provides an overview of FNT usage over the past few years (Section ~\ref{sec-review}). Conclusions derived from literature survey supports our HFNT$^\text{M}$ approach, where a Pareto-based multiobjective genetic programming was used for HFNT optimization (Section~\ref{sub-mo}). Section~\ref{sub-fnt} provides a detailed discussion on the basics of HFNT: MOGP for HFNT structure optimization, and DE for HFNT parameter optimization. The efficiency of the above-mentioned hybrid and complex multiobjective FNT algorithm (HFNT$^\text{M}$) was tested over various prediction problems using a comprehensive experimental set-up (Section~\ref{sec-exp}). The experimental results support the merits of proposed approach (Section~\ref{sec-res}). Finally, we provide a discussion of experimental outcomes in Section~\ref{sec-dis} followed by conclusions in Section~\ref{sec-con}.

\section{Literature Review}
\label{sec-review}
The literature survey describes the following points: basics of FNT, approaches that improvised FNT, and FNTs successful application to various real-life problems. Subsequently, the shortcomings of basic FNT version are concluded that inspired us to propose HFNT$ ^\text{M} $.

FNT was first proposed by Chen et al.~\cite{chen2005time}, where a tree-like-structure was optimized by using PIPE. Then, its approximation ability was tested for time-series forecasting~\cite{chen2005time} and intrusion detection~\cite{chen2005feature}, where a variant of simulated annealing (called degraded ceiling)~\cite{sanchez2001combining}, and particle swarm optimization (PSO)~\cite{kennedy2001swarm}, respectively, were used for FNT parameter optimization. Since FNT is capable of input feature selection, in~\cite{chen2006feature}, FNT was applied for selecting input features in several classification tasks, in which FNT structure was optimized by using genetic programming (GP)~\cite{riolo2014genetic}, and the parameter optimization was accomplished by using memetic algorithm~\cite{chen2011multi}. Additionally, they defined five different mutation operators, namely, changing one terminal node, all terminal nodes, growing a randomly selected sub-tree, pruning a randomly selected sub-tree, and pruning redundant terminals. Li et al.~\cite{li2006flexible} proposed FNT-based construction of decision trees whose nodes were conditionally replaced by neural node (activation node) to deal with continuous attributes when solving classification tasks. In many other FNT based approaches, like in~\cite{Chen2006137}, GP was applied to evolve hierarchical radial-basis-function network model, and in~\cite{Chen2007373} a multi-input-multi-output FNT model was evolved. Wu et al.~\cite{wu2007grammar} proposed to use grammar guided GP~\cite{Shan2004478} for FNT structure optimization. Similarly, in~\cite{jia2008mep}, authors proposed to apply multi-expression programming (MEP)~\cite{oltean2003evolving} for FNT structure optimization and immune programming algorithm~\cite{musilek2006immune} for the parameter vector optimization. To improve classification accuracy of FNT, Yang et al.~\cite{Yang2010690} proposed a hybridization of FNT with a further-division-of-partition-space method. In~\cite{bouaziz2016evolving}, authors illustrated crossover and mutation operators for evolving FNT using GP and optimized the tree parameters using PSO algorithm. 

A model is considered efficient if it has generalization ability. We know that a consensus decision is better than an individual decision. Hence, an ensemble of FNTs may lead to a better-generalized performance than a single FNT. To address this, in~\cite{Chen2007697}, authors proposed to make an ensemble of FNTs to predict the chaotic behavior of stock market indices. Similarly, in~\cite{yang2013ensemble}, the proposed FNTs ensemble predicted the breast cancer and network traffic better than individual FNT. In~\cite{ojha2016ensemble}, protein dissolution prediction was easier using ensemble than the individual FNT. 

To improve the efficiency in terms of computation, Peng et al.~\cite{peng2011parallel} proposed a parallel evolving algorithm for FNT, where the parallelization took place in both tree-structure and parameter vector populations. In another parallel approach, Wang et al.~\cite{wang2012modeling} used gene expression programming (GEP)~\cite{ferreira2006gene} for evolving FNT and used PSO for parameter optimization.

A multi-agent system~\cite{weiss1999multiagent} based FNT (MAS-FNT) algorithm was proposed in~\cite{ammar2015negotiation}, which used GEP and PSO for the structure and parameter optimization, respectively. The MAS-FNT algorithm relied on the division of the main population into sub-population, where each sub-population offered local solutions and the best local solution was picked-up by analyzing tree complexity and accuracy. 

Chen et al.~\cite{chen2005time,chen2006feature} referred the arbitrary choice of activation function at non-leaf nodes. However, they were restricted to use only Gaussian functions. A performance analysis of various activation function is available in~\cite{burianekperformance}. Bouaziz et al.~\cite{bouaziz2013hybrid,bouaziz2014universal} proposed to use beta-basis function at non-leaf nodes of an FNT. Since beta-basis function has several controlling parameters such as shape, size, and center, they claimed that the beta-basis function has advantages over other two parametric activation functions. Similarly, many other forms of neural tree formation such as balanced neural tree~\cite{micheloni2012balanced}, generalized neural tree~\cite{foresti2002generalized}, and convex objective function neural tree~\cite{rani2015neural}, were focused on the tree improvement of neural nodes. 

FNT was chosen over the conventional neural network based models for various real-world applications related to prediction modeling, pattern recognition, feature selection, etc. Some examples of such applications are cement-decomposing-furnace production-process modeling~\cite{shou2008modeling}, time-series prediction from gene expression profiling~\cite{yang2013reverse}. stock-index modeling~\cite{Chen2007697}, anomaly detection in peer-to-peer traffic~\cite{Chen2009685}, intrusion detection~\cite{novosad2010fast}, face identification~\cite{pan2007face}, gesture recognition~\cite{Guo20121099}, shareholder's management risk prediction~\cite{qu2011controlling}, cancer classification~\cite{rajini2012swarm}, somatic mutation, risk prediction in grid computing~\cite{abdelwahab2016ensemble}, etc. 

The following conclusions can be drawn from the literature survey. First, FNT was successfully used in various real-world applications with better performance than other existing function approximation models. However, it was mostly used in time-series analysis. Second, the lowest approximation error obtained by an individual FNT during an evolutionary phase was considered as the best structure that propagated to the parameter optimization phase. Hence, there was no consideration as far as structural simplicity and generalization ability are concerned. Third, the computational nodes of the FNT were fixed initially, and little efforts were made to allow for its automatic adaptation. Fourth, little attention was paid to the statistical validation of FNT model, e.g., mostly the single best model was presented as the experimental outcome. However, the evolutionary process and the meta-heuristics being stochastic in nature, statistical validation is inevitably crucial for performance comparisons. Finally, to create a generalized model, an ensemble of FNTs were used. However, FNTs were created separately for making the ensemble. Due to stochastic nature of the evolutionary process, FNT can be structurally distinct when created at different instances. Therefore, no explicit attention was paid to create diverse FNTs within a population itself for making ensemble. In this article, a heterogeneous FNT called HFNT was proposed to improve the basic FNT model and its performance by addressing above mentioned shortcomings.

\section{Multi-objectives and Flexible Neural Tree}
\label{sec-moFNT}
In this section, first, Pareto-based multiobjective is discussed. Second, we offer a detailed discussion on FNT and its structure and parameter optimization using NSGA-II-based MOGP and DE, respectively. Followed by a discussion on making an evolutionary weighted ensemble of the candidates from the final population.
\subsection{Pareto-Based Multi-objectives}
\label{sub-mo}
Usually, learning algorithms owns a single objective, i.e., the approximation error minimization, which is often achieved by minimizing mean squared error (MSE) on the learning data. MSE $ E $ on a learning data is computed as:
\begin{equation}
\label{eq_mse}
E =  \frac{1}{N} \sum\limits_{i = 1}^{N} (d_i - y_i)^2,
\end{equation} 
where $ d_i $ and $ y_i $ are the desired output and the model's output, respectively and $ N $ indicates total data pairs in the learning set. Additionally, a statistical goodness measure, called, correlation coefficient $ r $ that tells the relationship between two variables (i.e., between the desired output $ \mathbf{d} $ and the model's output $ \mathbf{y} $) may also be used as an objective. Correlation coefficient $ r $ is computed as:	 
\begin{equation}
\label{eq_corr}
r = \frac{\sum_{i = 1}^{N}\left(d_i - \mathbf{\bar{d_i}} \right) \left(y_i - \mathbf{\bar{y_i}} \right) }{ \sqrt{\sum_{i = 1}^{N}\left(d_i - \mathbf{\bar{d_i}} \right)^2 \sum_{i = 1}^{N}\left(y_i - \mathbf{\bar{y_i}} \right)^2}},
\end{equation} 
where  $ \mathbf{\bar{d}} $ and $ \mathbf{\bar{y}} $ are means of the desired output $ \mathbf{d} $ and the model's output $ \mathbf{y} $, respectively. 

However, single objective comes at the expense of model's complexity or generalization ability on unseen data, where generalization ability broadly depends on the model's complexity~\cite{jin2005evolutionary}. A common model complexity indicator is the number of free parameters in the model. The \emph{approximation error} \eqref{eq_mse} and the number of free parameters minimization are two conflicting objectives. One approach is to combine these two objectives as:
\begin{equation}
	\label{eq_scalarObj}
	f = \alpha E + (1 -\alpha ) D,
\end{equation} 
where $ 0 \le \alpha \le 1$  is a constant, $ E $ is the MSE~\eqref{eq_mse} and $ D $ is the total free parameter in a model. The scalarized objective $ f $ in~\eqref{eq_scalarObj}, however, has two disadvantages. First, determining an appropriate $ \alpha $ that controls the conflicting objectives. Hence, generalization ability of the produced model will be a mystery~\cite{das1997closer}. Second, the scalarized objective $ f $ in~\eqref{eq_scalarObj} leads to a single best model that tells nothing about how the conflicting objectives were achieved. In other words, no single solution exists that may satisfy both objectives, simultaneously. 

We study a multiobjective optimization problem of the form:~\\
\noindent $\mbox{minimize } \{f_1(\mathrm{\mathbf{w}}), f_2(\mathrm{\mathbf{w}}), \ldots, f_m(\mathrm{\mathbf{w}})\}$\\
\noindent $\mbox{subject to } \mathrm{\mathbf{w}} \in W  \nonumber $\\
\noindent where we have $m \ge 2$ objective functions $f_i: \mathbb{R}^n \to  \mathbb{R}$. We denote the vector 
of objective functions by $ \text{\textbf{f}}(\mathrm{\mathbf{w}}) = \langle f_1(\mathrm{\mathbf{w}}), f_2(\mathrm{\mathbf{w}}), \ldots, f_m(\mathrm{\mathbf{w}})\rangle$. The decision (variable) vectors $\mathrm{\mathbf{w}} = \langle w_1, w_2,\ldots, w_n\rangle$
belong to the set $W \subset  \mathbb{R}^n$, which is a subset of the decision variable space $ \mathbb{R}^n$. 
The word `minimize' means that we want to minimize all the objective functions simultaneously. 

A nondominated solution is one in which no one objective function can be improved without a simultaneous detriment to at least one of the other objectives of the solution~\cite{deb2000fast}. The nondominated solution is also known as a Pareto-optimal solution.
\begin{mydef}
	Pareto-dominance - A solution  $ \mathrm{\mathbf{w}}_1 $ is said to dominate a solution $ \mathrm{\mathbf{w}}_2 $ if $ \forall i = 1,2,\ldots,m, $ $ f_i(\mathrm{\mathbf{w}}_1) \le f_i(\mathrm{\mathbf{w}}_2) $, and there exists $ j \in \left\lbrace 1,2,\ldots,m\right\rbrace  $ such that $ f_j(\mathrm{\mathbf{w}}_1) < f_j(\mathrm{\mathbf{w}}_2) $ holds.
\end{mydef}
\begin{mydef}
	Pareto-optimal - A solution $ \mathrm{\mathbf{w}}_1 $ is called \textit{Pareto-optimal} if there does not exist any other solution that dominates it. A set \textit{Pareto-optimal} solution is called \textit{Pareto-front}.   
\end{mydef}
Algorithm \ref{algo_nsga} is a basic framework of NSGA-II based MOGP, which was used for computing Pareto-optimal solutions from an initial HFNT population. The individuals in MOGP were sorted according to their dominance in population. Note that the function $ size(\cdot) $ returns total number of rows (population size) for a 2-D matrix and returns total number of elements for a vector. The Moreover, individuals were sorted according to the rank/Pareto-front. MOGP is an elitist algorithm that allowed the best individuals to propagate into next generation. Diversity in the population was maintained by measuring crowding distance among the individuals~\cite{deb2000fast}.

~\\
\begin{algorithm}[H]
	\KwData{Problem and Objectives}
	\KwResult{A bag $ M $ of solutions selected from Pareto-fronts}
	initialization: HFNT population $ P $\;
	evaluation: nondominated sorting of $ P $ \;	
	\While{termination criteria not satisfied}{
		selection: binary tournament selection\;
		generation: a new population $ Q $\;
		recombination: $ R = P + Q $\;
		evaluation: nondominated sorting of $ R $\;	
		elitism: $ P$ =  $ size(P) $ best individuals from $ R $\;
	}
	\caption{NSGA-II based multiobjective genetic programming}
    \label{algo_nsga}
\end{algorithm}

\subsection{Heterogeneous Flexible Neural Tree}
\label{sub-fnt}
HFNT is analogous to a multi-layer feedforward neural network that has over-layer connections and activation function at the nodes. HFNT construction has two phases~\cite{chen2005time}: 1) the \emph{tree construction} phase, in which evolutionary algorithms are applied to construct tree-like structure; and 2) the \emph{parameter-tuning} phase, in which genotype of HFNT (underlying parameters of tree-structure) is optimized by using parameter optimization algorithms. 

\begin{figure}
	\centering
	\includegraphics[width=0.6\textwidth]{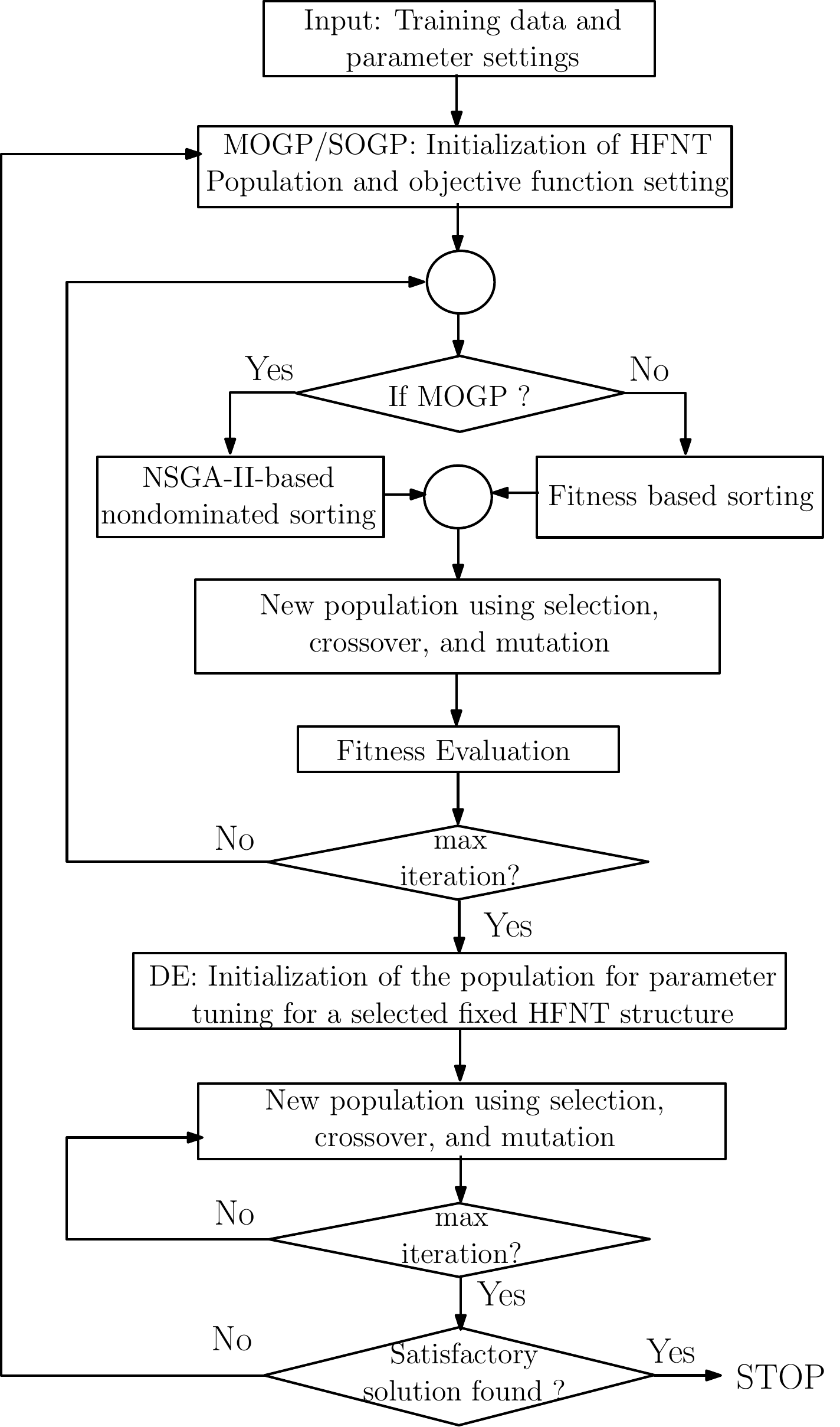}
	\caption{Co-evolutionary construction of the heterogeneous flexible neural tree.}
	\label{fig_FNT_dia}
\end{figure}

To create a near-optimum model, phase one starts with random tree-like structures (population of initial solutions), where parameters of each tree are fixed by a random guess. Once a near-optimum tree structure is obtained, \emph{parameter-tuning} phase optimizes its parameter. The phases are repeated until a satisfactory solution is obtained. Figure~\ref{fig_FNT_dia} is a lucid illustration of these two phases that work in some co-evolutionary manner.  From Figure~\ref{fig_FNT_dia}, it may be observed that two global search algorithms MOGP (for structure optimization) and DE (for parameter optimization) works in a nested manner to obtain a near optimum tree that may have less complex tree structure and better parameter. Moreover, evolutionary algorithm allowed HFNT to select activation functions and input feature at the nodes from sets of activation functions and input features, respectively. Thus, HFNT possesses automatic feature selection ability.

\subsubsection{Basic Idea of HFNT}	
\begin{figure}	
	\centering
	\includegraphics[scale = 0.7]{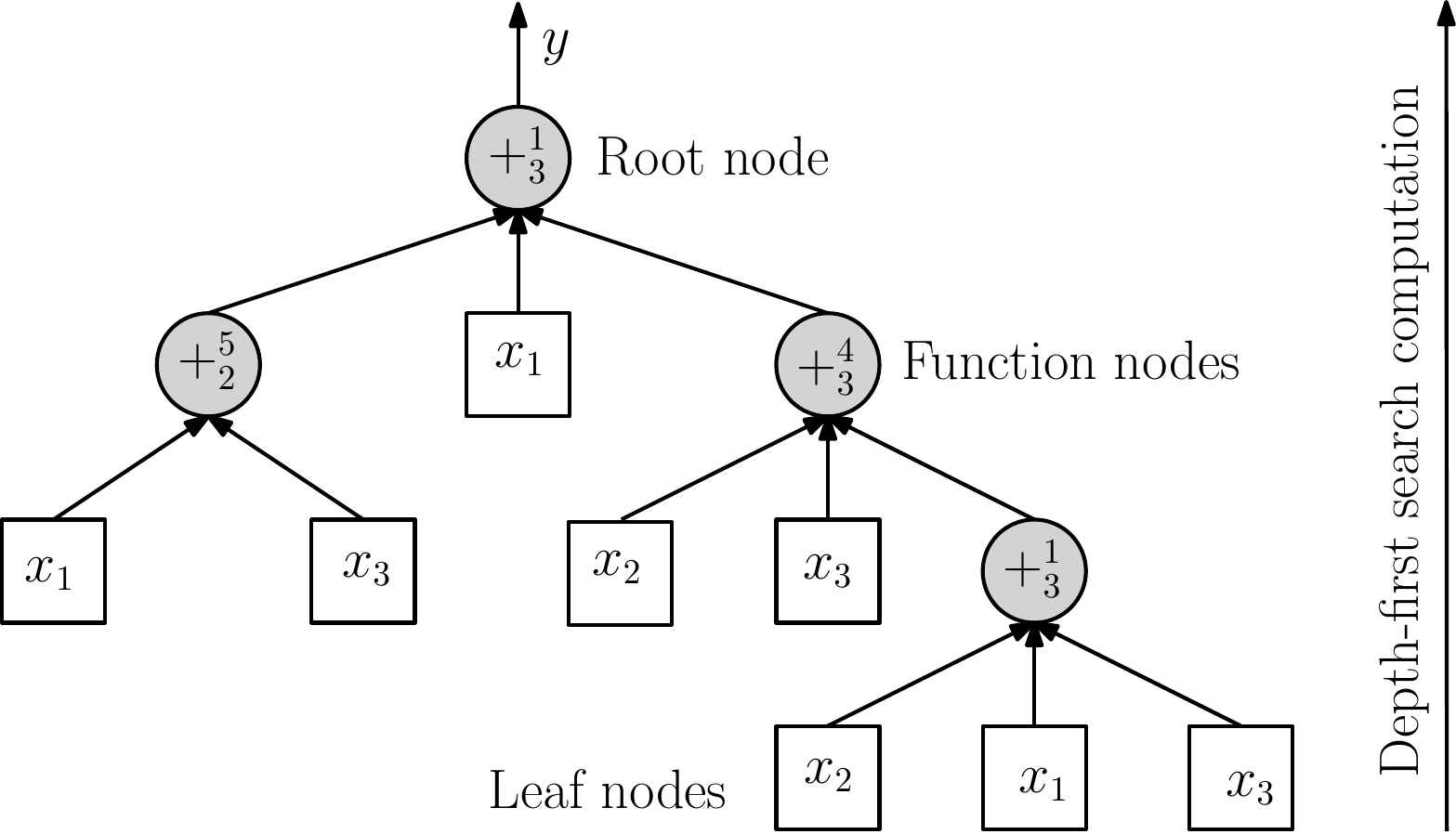}
	\caption{Typical representation of a neural tree $  S = F \cup T$ whose function instruction set $ F = \left\lbrace +^1_3, +^4_2, +^5_3\right\rbrace  $ and terminal instruction set $T = \left\lbrace x_1, x_2, x_3,x_4\right\rbrace  $.}
	\label{fig_fntGen}
\end{figure}

\begin{figure}	
	\centering
    \includegraphics[scale = 1.5]{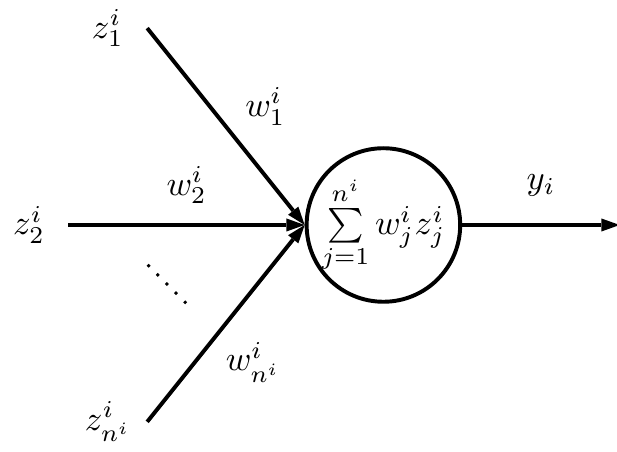}
	\caption{ Illustration of a computational node. The variable $ n^i $ indicates the number of inputs $ z^i_j $ and weights $ w^i_j $ received at the $ i $-th node and the variable  $ y^i $ is the output of the $ i $-th node.}
	\label{fig_fntGen_node}
\end{figure}

An HFNT $ S $ is a collection of function set $ F $ and instruction set $ T $:
\begin{equation}
	\label{eq-fnt}
	S = F \cup T = \left\lbrace +^{U(k)}_2,+^{U(k)}_3,\cdots,+^{U(k)}_{tn}  \right\rbrace  \cup \left\lbrace x_1,x_2,\ldots,x_d \right\rbrace 
\end{equation}
where $ +^k_j \left( j = 2,3,\ldots,tn \right)  $ denotes a non-leaf instruction (a computational node). It receives $  2 \le j \le tn $ arguments and $ U(k) $ is a function that randomly takes an activation function from a set of $ k $ activation functions. Maximum arguments  $ tn $ to a computational node are predefined. A set of seven activation functions is shown in Table~\ref{tab_actFun}. Leaf node's instruction $ x_1,x_2,\ldots,x_d $ denotes input variables. Figure~\ref{fig_fntGen} is an illustration of a typical HFNT. Similarly, Figure~\ref{fig_fntGen_node} is an illustration of a typical node in an HFNT. 

The $i$-th computational node (Figure~\ref{fig_fntGen_node}) of a tree (say $ i $-th node in Figure~\ref{fig_fntGen}) receives $ n^i $ inputs (denoted as $ z^i_j $) through $ n^i $ connection-weights (denoted as $ w^i_j $) and takes two adjustable parameters $ a^i $ and $ b^i $ that represents the arguments of the activation function $ \varphi^k_i(.) $ at that node. The purpose of using an activation function at a computational node is to limit the output of the computational node within a certain range. For example, if the $i$-th node contains a Gaussian function $ k=1 $ (Table~\ref{tab_actFun}). Then, its output $ y_i $ is computed as:
\begin{equation}
\label{eq_netNode}
y_i= \varphi^k_i (a_i, b_i, o_i )=\exp \left(-\left(\frac{o_i - a_i}{b_i}\right)\right)
\end{equation}
where $ o_i $ is the weighted summation of the inputs $ z^i_j$ and weights $ w_j^i $ $ (j= 1 \text{ to }n^i)$ at the $i$-th computational node (Figure~\ref{fig_fntGen_node}), also known as excitation of the node. The net excitation $ o^i $ of the $ i $-th node is computed as:
\begin{equation}
\label{eq-outNode}
o_i= \sum\limits_{j=1 }^{n^i}  w_j^i z_j^i 
\end{equation} 
where $ z_j^i \in \{x_1,x_2, \ldots ,x_d\} $ or, $ z_j^i \in \{y_1,y_2, \ldots,y_m\} $, i.e., $ z_j^i $ can be either an input feature (leaf node value) or the output of another node (a computational node output) in the tree. Weight $ w_j^i $ is the connection weight of real value in the range $[w_l,w_u]$. Similarly, the output of a tree $ y $ is computed from the root node of the tree, which is recursively computed by computing each node's output using~\eqref{eq_netNode} from right to left in a depth-first method.

The fitness of a tree depends on the problem. Usually, learning algorithm uses \emph{approximation error}, i.e., MSE \eqref{eq_mse}. Other fitness measures associated with the tree are \emph{tree size} and \emph{diversity index}. The \emph{tree size} is the number of nodes (excluding root node) in a tree, e.g., the number of computational nodes and leaf nodes in the tree in Figure~\ref{fig_fntGen} is 11 (three computational nodes and eight leaf-nodes). The number of distinct activation functions (including root node function) randomly selected from a set of activation functions gives the \emph{diversity index} of a tree. Total activation functions (denoted as $ k $ in $ +^k_j $) selected by the tree in Figure~\ref{fig_fntGen} is three ($ +^1_3, +^4_3, \text{ and } +^5_3 $). Hence, its \emph{diversity index} is three.
\begin{table}
	\centering
	\caption{Set of activation function used in neural tree construction}
	\footnotesize 
	\label{tab_actFun}
	\begin{tabular}{lll}
		\hline
		Activation-function  & $ k $ & Expression for $ \varphi^k_i (a,b,x ) $\\ 
		\hline
		Gaussian Function              & 1  &  $ f(x,a,b) = \exp\left( -((x-a)^2)/(b^2)  \right)   $\\
		Tangent-Hyperbolic            & 2 &  $ f(x) = (e^x - e^{-x})/(e^x + e^{-x})    $\\
		Fermi Function                    & 3 &  $ f(x) = 1/(1 + e^{-x})    $\\
		Linear Fermi                        & 4 &  $ f(x,a,b) = a \times 1/((1 + e^{-x})) + b   $\\
		Linear Tangent-hyperbolic & 5 &  $ f(x,a,b) =a \times (e^x - e^{-x})/(e^x + e^{-x}) + b    $\\
		Bipolar Sigmoid                   & 6 &  $ f(x,a) = (1 - e^{-2xa})/(a(1 + e^{-2xa})) $\\
		Unipolar Sigmoid                 & 7 &  $ f(x,a) = (2  |a|)/(1 + e^{-2|a|x})   $\\		
		\hline
	\end{tabular}
\end{table}

\subsection{Structure and Parameter Learning (Near optimal Tree)}
A tree that offers the lowest \emph{approximation error} and the simplest structure is a near optimal tree, which can be obtained by using an evolutionary algorithm such as GP~\cite{riolo2014genetic}, PIPE~\cite{salustowicz1997probabilistic}, GEP~\cite{ferreira2006gene}, MEP~\cite{oltean2003evolving}, and so on. To optimize tree parameters, algorithms such as genetic algorithm~\cite{eiben2015ec}, evolution strategy~\cite{eiben2015ec}, artificial bee colony~\cite{karaboga2007powerful}, PSO~\cite{kennedy2001swarm,zhang2015comprehensive}, DE~\cite{das2016recent}, and any hybrid algorithm such as GA and PSO~\cite{juang2004hybrid} can be used. 

\subsubsection{Tree-construction}
The proposed multiobjective optimization of FNT has three fitness measures: \emph{approximation error}~\eqref{eq_mse} minimization, \emph{tree size} minimization, and \emph{diversity index} maximization. These objectives are simultaneously optimized during the \emph{tree construction} phase using MOGP, which guides an initial population $ P $ of random tree-structures according to Algorithm~\ref{algo_nsga}. The detailed description of the components of Algorithm~\ref{algo_nsga} are as follows:

\paragraph{Selection} In selection operation, a \emph{mating pool} of size $ size(P)r $ is created using \emph{binary tournament selection}, where two candidates are randomly selected from a population and the best (according to rank and crowding distance) among them is placed into  the mating pool. This process is continued until the mating pool is full.  An offspring population $ Q $ is generated by using the individuals of mating pool. Two distinct individuals (parents) are randomly selected from the mating pool to create new individuals using genetic operators: crossover and mutation. The crossover and mutation operators are applied with probabilities $ pc $ and $ pm $, respectively.

\paragraph{Crossover} In crossover operation, randomly selected sub-trees of two parent trees were swapped. The swapping includes the exchange of activation-nodes, weights, and inputs as it is described in~\cite{bouaziz2016evolving,eiben2015ec,Wongseree2007771}.

\paragraph{Mutation} The mutation of a selected individual from mating pool took place in the following manner~\cite{bouaziz2016evolving,eiben2015ec,Wongseree2007771}:
\begin{enumerate}[1)]
	\item A randomly selected terminal node is replaced by a newly generated terminal node.
	\item All terminal nodes of the selected tree were replaced by randomly generated new terminal nodes.
	\item A randomly selected terminal node or a computational node is replaced by a randomly generated sub-tree.
	\item A randomly selected terminal node is replaced by a randomly generated computational node. 
\end{enumerate}
In the proposed MOGP, during the each mutation operation event, one of the above-mentioned four mutation operators was randomly selected for mutation of the tree.
\paragraph{Recombination} The offspring population $ Q $ and the main population $ P $, are merged to make a combined population $ R $.
\paragraph{Elitism} In this step, $ size(Q) $ worst individuals are weeded out. In other words, $ size(P) $ best individuals are propagated to a new generation as main population $ P $.

\subsubsection{Parameter-tuning} 
In \emph{parameter-tuning} phase, a single objective, i.e., \emph{approximation error} was used in optimization of HFNT parameter by DE. The tree parameters such as weights of tree edges and arguments of activation functions were encoded into a vector $ \text{\textbf{w}} = \langle w_1, w_2,\ldots, w_n \rangle$ for the optimization. In addition, a cross-validation (CV) phase was used for statistical validation of HFNTs.

The basics of DE is as follows. For an initial population $ H $ of parameter vectors $ \text{\textbf{w}}_i \text{ for } i = 1 \text{ to } size(H)$, DE repeats its steps mutation, recombination, and selection until an optimum parameter vector $ \mathrm{\mathbf{w}}^*$ is obtained. DE updates each parameter vector $ \mathrm{\mathbf{w}}_i \in H$ by selecting the best vector $ \text{\textbf{w}}^g_i  $ and three random vectors $ \mathrm{\mathbf{r}}^0_i, \mathrm{\mathbf{r}}^1_i,$ and $ \mathrm{\mathbf{r}}^2_i $ from $ H $ such that $ \mathrm{\mathbf{r}}^0_i \ne \mathrm{\mathbf{r}}^1_i \ne \mathrm{\mathbf{r}}^2_i $ holds. The random vector  $ \mathrm{\mathbf{r}}^0 $ is considered as a trial vector $ \mathrm{\mathbf{w}}^t_i $. Hence, for all $ i = 1, 2,\ldots, size(H)$, and $ j = 1,2,\ldots,n $, the $ j $-th variable $ w^t_{ij} $ of $ i $-th trail-vectors $ \mathrm{\mathbf{w}}^t_i $ is generated by using \emph{crossover}, \emph{mutation}, and  \emph{recombination} as:

\begin{equation}
\label{eq:deMutation}
w^t_{ij} = \left\lbrace 
\begin{array}{ll}
r^{(0)}_{ij} +  F (w^g_{ij} - r^0_{ij}) + F (r^1_{ij} - r^2_{ij}) &  u_{ij} < cr \parallel  j = k\\
r^{(0)}_{ij} &  u_{ij} \ge cr  \text{ \& } j \ne k\\
\end{array}	
\right.
\end{equation}
where $ k $ is a random index in $ [1,n]$, $ u_{ij}$ is within $ [0,1] $, $ k $ is in $ \left\lbrace 1,2,\ldots, n\right\rbrace  $, $ cr $ is crossover probability, and  $ F \in [0,2] $ is mutation factor. The trail vector $ \mathrm{\mathbf{w}}^t_i $ is selected if

\begin{equation}
\label{eq:deRecombination}
\mathrm{\mathbf{w}}_i = \left\lbrace 
\begin{array}{ll}
\mathrm{\mathbf{w}}^t_i  &  f(\mathrm{\mathbf{w}}^t_i ) < f(\mathrm{\mathbf{w}}_i ) \\
\mathrm{\mathbf{w}}_i &  f(\mathrm{\mathbf{w}}^t_i ) \ge f(\mathrm{\mathbf{w}}_i )\\
\end{array}		
\right.
\end{equation}
where $ f(.) $ returns fitness of a vector as per~\eqref{eq_mse}. Hence, the process of \emph{crossover}, \emph{mutation}, \emph{recombination}, and \emph{selection} are repeated until an optimal parameter vector solution $ \mathrm{\mathbf{w}}^* $ is found.

\subsection{Ensemble: Making use of MOGP Final Population} 
In \emph{tree construction} phase, MOGP provides a population from which we can select tree models for making the ensemble. Three conflicting objectives such as \emph{approximation error}, \emph{tree size}, and \emph{diversity index} allows the creation of Pareto-optimal solutions, where solutions are distributed on various Pareto-optimal fronts according to their rank in population. Ensemble candidates can be selected from the first line of solutions (Front 1), or they can be chosen by examining the three objectives depending on the user's need and preference. Accuracy and diversity among the ensemble candidate are important~\cite{kuncheva2003measures}. Hence, in this work, \emph{approximation error}, and diversity among the candidates were given preference over \emph{tree size}. Not to confuse ``\emph{diversity index}'' with ``diversity''.  The \emph{diversity index} is an objective in MOGP, and the diversity is the number of distinct candidates in an ensemble.  A collection $ M $ of the diverse candidate is called a bag of candidates~\cite{hastie2009elements}. In this work, any two trees were considered diverse (distinct) if the followings hold: 1) Two trees were of different size. 2) The number of function nodes/or leaf nodes in two trees were dissimilar. 3) Two models used a different set of input features. 4) Two models used a different set of activation functions.  Hence, diversity $ div $ of ensemble $ M $ (a bag of solutions) was computed as:
\begin{equation}
\label{eq-diversity}
div = \dfrac{distinct(M)}{size(M)},
\end{equation}         
where $ distinct(M) $ is a function that returns total distinct models in an ensemble $ M $ and $ size(M) $ is a total number of models in the bag.  

Now, for a classification problem, to compute combined vote of respective candidate's outputs $ m_1$, $m_2$, $\ldots$, $m_{size(M)} $ of bag $ M $ and classes $ \omega_1,\omega_2,\ldots, \omega_C$, we used an indicator function  $ \mathbb{I}\left( . \right)  $ which takes $ 1 $ if `$ . $' is true, and takes $ 0 $ if `$ . $' is false. Thus, ensemble decisions by weighted majority voting is computed as~\cite{polikar2006ensemble,zhou2012ensemble}: 
\begin{equation}
\label{eq_ensmbleWMV}
y =  \arg\max_{j = 1}^C \sum\limits_{t = 1}^{size(M)} w_t \mathbb{I}\left( m_t = \omega_j \right) ,   
\end{equation}     
where $ w_t $ is weight associated with the $ t $-th candidate $ m_t $ in an ensemble $ M $ and $ y $ is set to class $ \omega_j $ if the total weighted vote received by $ \omega_j $ is higher than the total vote received by any other class.  Similarly, the ensemble of regression methods was computed by weighted arithmetic mean as~\cite{polikar2006ensemble}:
\begin{equation}
\label{eq_ensmbleWAM}
y =  \sum\limits_{t = 1}^{size(M)} w_t \, m_t,   
\end{equation}  
where $ w_t $ and $ m_t $ are weight and output of $ t $-th candidate in a bag $ M $, respectively, and $ y $ is the ensemble output, which is then used for computing MSE~\eqref{eq_mse} and correlation coefficient~\eqref{eq_corr}. The weights may be computed according to fitness of the models, or by using a metaheuristic algorithm. In this work, DE was applied to compute the ensemble weights $ w_t $, where population size was set to 100 and number of function evaluation was set to 300,000. 

\subsection{Multiobjective: A General Optimization Strategy}
\label{subsub_genAlgo}
A summary of general HFNT learning algorithm is as follows:
\begin{enumerate}[ Step 1.]
	\item Initializing HFNT training parameters.
	\item Apply \emph{tree construction} phase to guide initial HFNT population towards Pareto-optimal solutions.
	\item Select tree-model(s) from MOGP final population according to their \emph{approximation error}, \emph{tree size}, and \emph{diversity index} from the Pareto front.
	\item Apply \emph{parameter-tuning} phase to optimize the selected tree-model(s).
	\item Go to Step 2, if no satisfactory solution found. Else go to Step 6.
	\item Using a cross-validation (CV) method to validate the chosen model(s).
	\item Use the chosen tree-model(s) for making ensemble (recommended).
	\item Compute ensemble results of the ensemble model (recommended).	
\end{enumerate}

\section{Experimental Set-Up}
\label{sec-exp}
Several experiments were designed for evaluating the proposed HFNT$^\text{M}$. A careful parameter-setting was used for testing its efficiency. A detailed description of the parameter-setting is given in Table~\ref{tab_fntParameters}, which includes: definitions, default range, and selected value. The phases of the algorithm were repeated until the stopping criteria met, i.e., either the lowest predefined \emph{approximation error} was achieved, or the maximum function evaluations were reached. The repetition holds the key to obtaining a good solution. A carefully designed repetition of these two phases may offer a good solution in fewer of function evaluations.

In this experiment, three general repetitions $ i_g $ were used with 30 \emph{tree construction} iterations $ i_s $, and 1000 \emph{parameter-tuning} iterations $ i_p $ (Figure~\ref{fig_FNT_dia}). Hence, the maximum function evaluation\footnote{Initial GP population + three repetition ((GP population + mating pool size) $ \times $ MOGP iterations + MH population $ \times $ MH iterations) = $ 30 + 3 \times [(30 + 15) \times 30 + 50 \times 1000] = 154,080 $.} $[size(P)+i_g\{i_s(size(P)+size(P)r)+ i_psize(H)\}]$ was $ 154,080 $. The DE version $DE/rand-to-best/1/bin$~\cite{das2016recent} with $ cr $ equal to 0.9 and $ F $ equal to 0.7 was used in the \emph{parameter-tuning} phase. 

\begin{table}
	\centering
	\footnotesize 		
	\caption{Multiobjective flexible neural tree parameter set-up for the experiments}
	\label{tab_fntParameters}
	\setlength{\tabcolsep}{0.15cm}
	\begin{tabular}{llll}
		\hline
		Parameter & Definition & Default Rang &  Value \\
		\hline
		Scaling & Input-features scaling range. & $[dl,du],$ $ dl \in \mathbb{R}$,  $du \in \mathbb{R}$  & [0,1] \\	
		Tree height & Maximum depth (layers) of a tree model. & $ \left\lbrace td \in  \mathbb{Z}^+| td > 1 \right\rbrace $  & 4 \\			
		Tree arity  & Maximum arguments of a node $ +^k_{tn} $. & $ \left\lbrace tn \in  \mathbb{Z}^+| n \ge 2 \right\rbrace $  & 5 \\			
		Node range & Search space of functions arguments. & $[nl,nu],$ $ nl \in \mathbb{R}$,  $nu \in \mathbb{R}$  & [0,1] \\		
		Edge range & Search space for edges (weights) of tree. & $[w_l,w_u],$ $ w_l\in \mathbb{R}$,  $w_u \in \mathbb{R}$  & [-1,1] \\			
		$ P $  & MOGP population.  & $  size(P) > 20  $    & 30  \\			
		Mutation  & Mutation probability  & $ pm  $  & 0.3  \\		
		Crossover  & Crossover probability  & $ pc = 1 - pm $ & 0.7 \\		
		Mating pool  & Size of the pool of selected candidates. & $  size(P)r,$ $  0 \le r \le 1$  & 0.5 \\			
		Tournament  & Tournament selection size. &  $ 2 \le bt \le size(P)$   & 2 \\		
		$ H $ & DE population. & $ size(H) \ge 50  $   & 50 \\				
		General $ i_g $ & Maximum number of trails. &$ \left\lbrace i_g \in  \mathbb{Z}^+| i_g > 1 \right\rbrace $   & 3  \\
		Structure $ i_s $ &  MOGP iterations & $ \left\lbrace i_s \in  \mathbb{Z}^+| i_s \ge 50 \right\rbrace $    & 30  \\
		Parameter $ i_p $ &  DE iterations &$ \left\lbrace i_p \in  \mathbb{Z}^+| i_p \ge 100 \right\rbrace $    & 1000  \\
		\hline
	\end{tabular}
\end{table}	

The experiments were conducted over classification, regression, and time-series datasets. A detailed description of the  chosen dataset from the UCI machine learning~\cite{UCILichman2013} and KEEL~\cite{alcala2009keel} repository is available in Table~\ref{tab-dataset}. The parameter-setting mentioned in Table~\ref{tab_fntParameters} was used for the experiments over each dataset.  Since the stochastic algorithms depend on random initialization, a pseudorandom number generator called, Mersenne Twister algorithm that draws random values using probability distribution in a pseudo-random manner was used for initialization of HFNTs~\cite{matsumoto1998mersenne}.  Hence, each run of the experiment was conducted with a random seed drawn from the system. We compared HFNT$^\text{M}$ performance with various other approximation models collected from literature. A list of such models is provided in Table~\ref{tab_litModels}. A developed software tool based on the proposed HFNT$^\text{M}$ algorithm for predictive modeling is available in~\cite{ojha2016mogp}.

\begin{figure}
	\centering
	\subfigure[Error versus \emph{tree size} versus \emph{diversity index}]
	{
		\includegraphics[width = 0.46\textwidth]{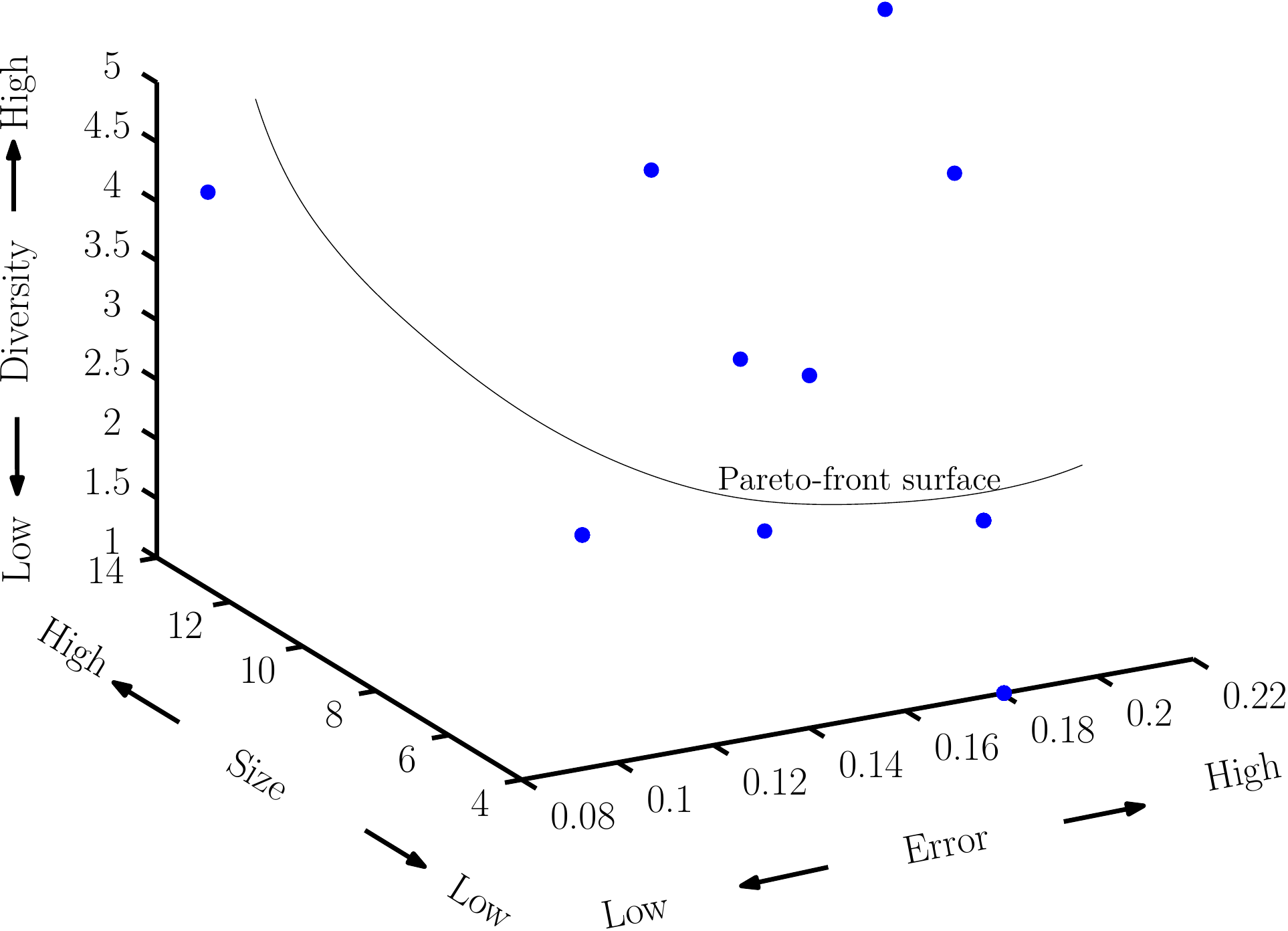}
		\label{fig_graph1}%
	}
	\subfigure[Error versus \emph{tree size} and \emph{diversity index}]
	{
		\includegraphics[width = 0.46\textwidth]{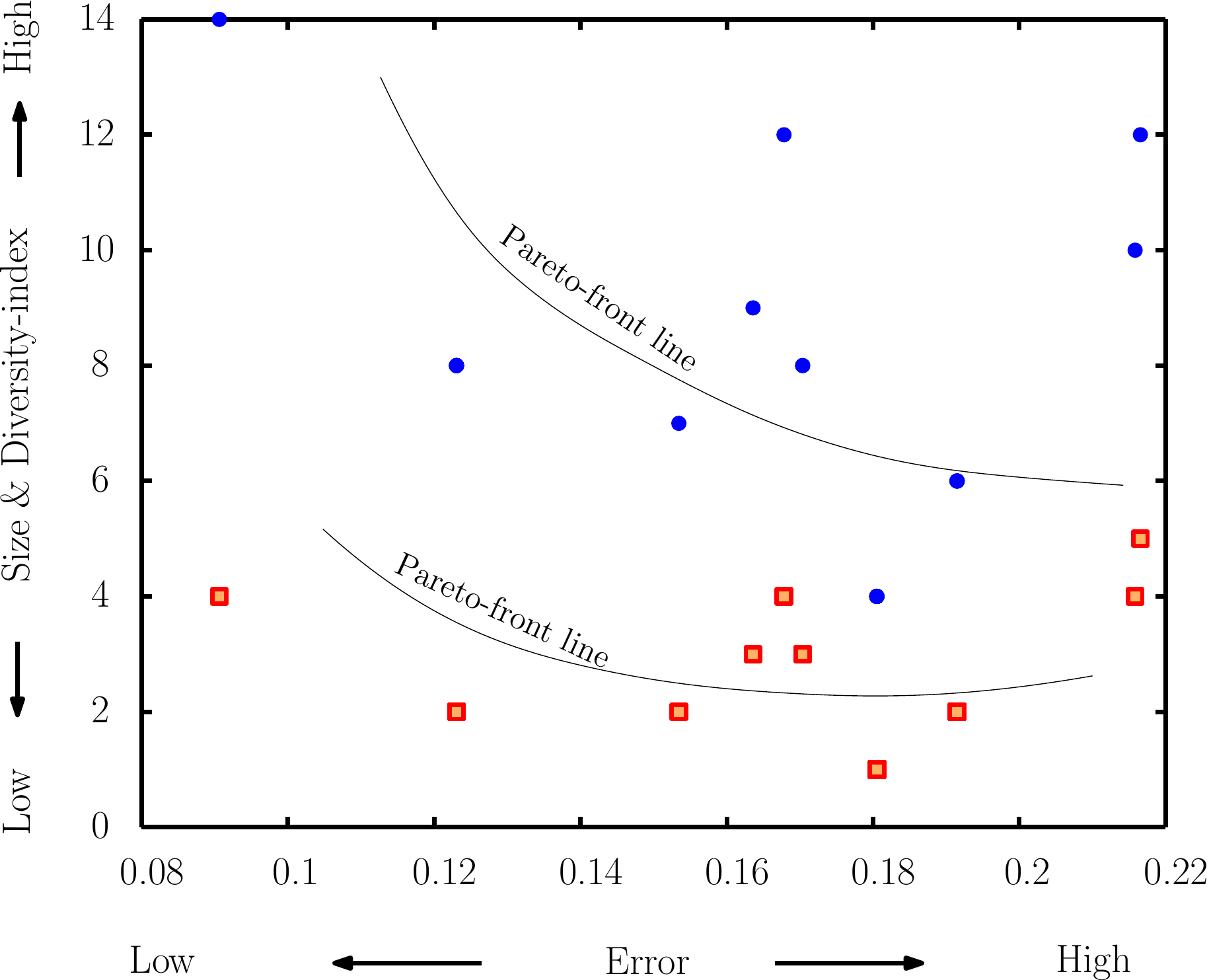}
		\label{fig_graph2}%
	}
	\caption{ Pareto-front of a final population of 50 individuals generated from the training dataset of time-series problem MGS. (a) 3-D plot of solutions and a Pareto-front is a surface. (b) 2-D plot of Error versus complexity (in blue dots) and Error versus diversity (in red squares).}
	\label{fig_paretoFront}
\end{figure}

To construct good ensemble systems, highly diverse and accurate candidates were selected in the ensemble bag $ M $. To increase diversity~\eqref{eq-diversity} among the candidates, the Pareto-optimal solutions were examined by giving preference to the candidates with low \emph{approximation error}, small \emph{tree size} and distinct from others selected candidates. Hence, $ size(M) $ candidates were selected from a population $ P $. An illustration of such selection method is shown in Figure~\ref{fig_paretoFront}, which represents an MOGP final population of 50 candidate solutions computed over dataset MGS.       

MOGP simultaneously optimized three objectives. Hence, the solutions were arranged on the three-dimensional map (Figure~\ref{fig_graph1}), in which along the x-axis, error was plotted; along the y-axis, \emph{tree size} was plotted; and along z-axis, \emph{diversity index} (diversity) was plotted. However, for the simplicity, we have arranged solutions also in 2-D plots (Figure~\ref{fig_graph2}), in which along the x-axis, computed error was plotted; and along the y-axis, \emph{tree size} (indicated by blue dots) and \emph{diversity index} (indicated by red squares) were plotted. From Figure~\ref{fig_graph2}, it is evident that a clear choice is difficult since decreasing \emph{approximation error} increases models \emph{tree size} (blue dots in Figure~\ref{fig_graph2}). Similarly, decreasing \emph{approximation error} increases models \emph{tree size} and diversity (red squares in Figure~\ref{fig_graph2}). Hence, solutions along the Pareto-front (rank-1), i.e., Pareto surface indicated in the 3-D map of the solutions in Figure~\ref{fig_graph1} were chosen for the ensemble. For all datasets, ensemble candidates were selected by examining Pareto-fronts in a similar fashion as described for the dataset MGS in Figure~\ref{fig_paretoFront}. 

\begin{figure}
	\centering
	\subfigure[Single objective optimization]
	{
		\includegraphics[width = 0.47\textwidth]{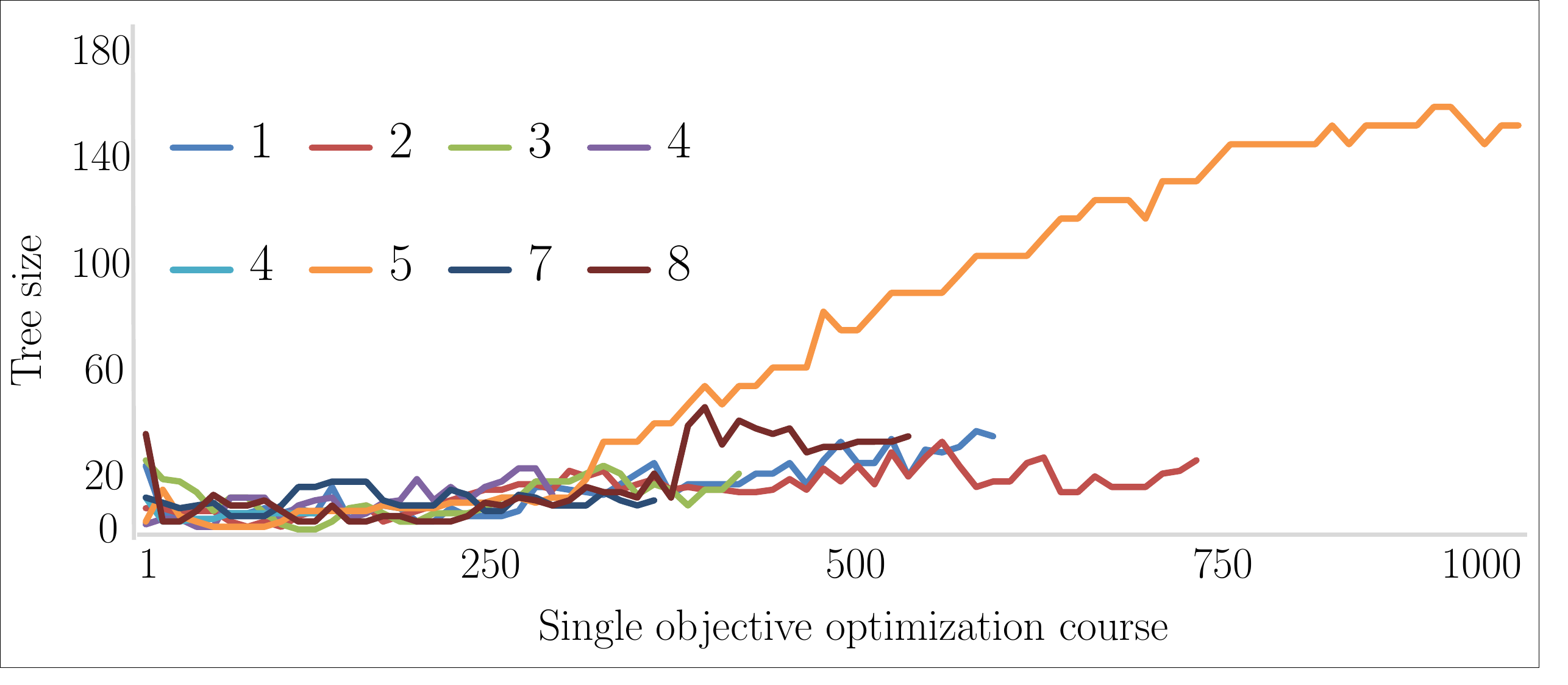} 
		\label{fig_so_graph}%
	}
	\subfigure[Multiobjective objective optimization ]
	{
		\includegraphics[width = 0.47\textwidth]{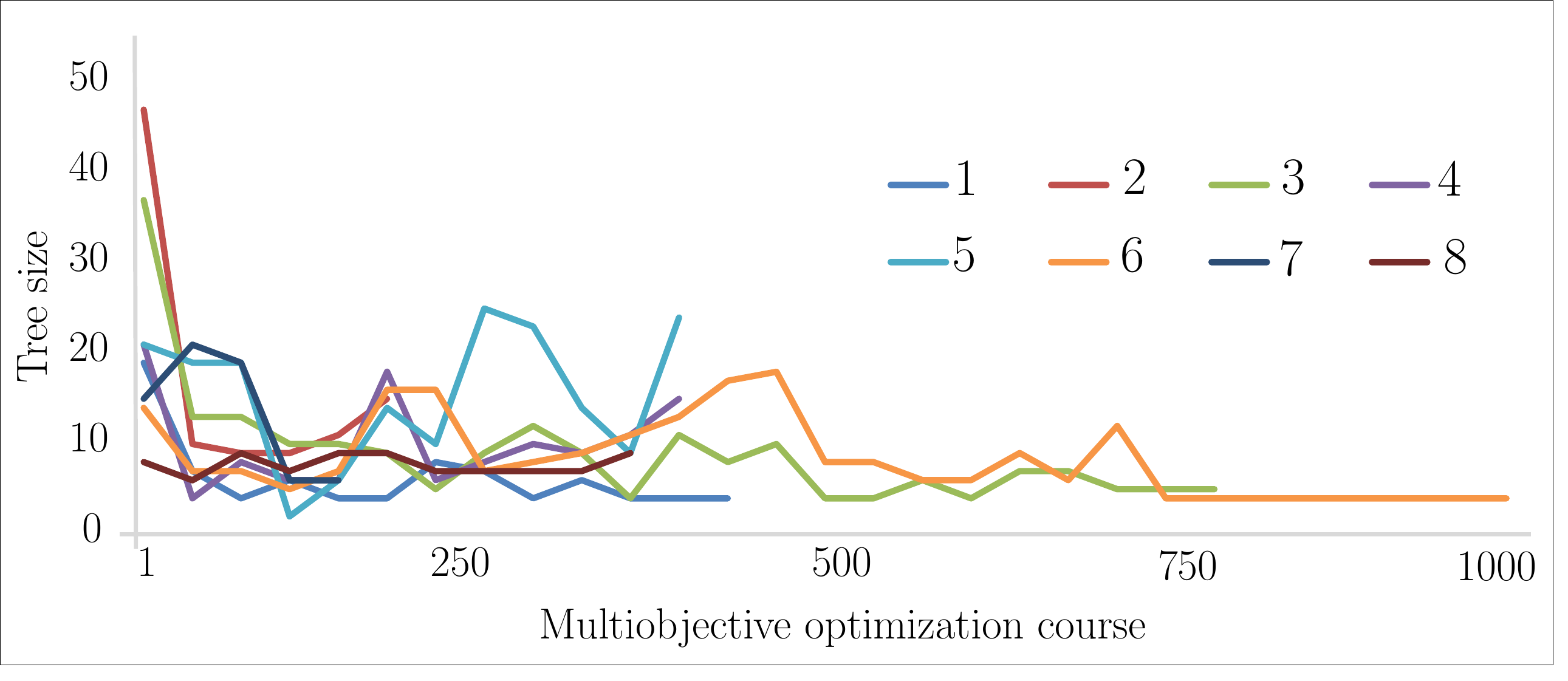}
		\label{fig_mo_graph}%
	}
	\caption{Comparison of single and multiobjective optimization course.}
	\label{fig_svmPlots}
\end{figure}

The purpose of our experiment was to obtain sufficiently good prediction models by enhancing predictability and lowering complexity. We used MOGP for optimization of HFNTs. Hence, we were compromising fitness by lowering models complexity. In single objective optimization, we only looked for models fitness. Therefore, we did not possess control over model's complexity. Figure~\ref{fig_svmPlots} illustrates eight runs of both single and multiobjective optimization course of HFNT, where models \emph{tree size} (complexity) is indicated along y-axis and x-axis indicates fitness value of the HFNT models. The results shown in Figure~\ref{fig_svmPlots} was conducted over MGS dataset. For each single objective GP and multiobjective GP, optimization course was noted, i.e., successive fitness reduction and \emph{tree size} were noted for 1000 iterations. 

It is evident from Figure~\ref{fig_svmPlots} that the HFNT$^\text{M}$ approach leads HFNT optimization by lowering model's complexity. Whereas, in the single objective, model's complexity was unbounded and was abruptly increased. The average \emph{tree size} of eight runs of single and eight runs of multiobjective were 39.265 and 10.25, respectively; whereas, the average fitness were 0.1423 and 0.1393, respectively. However, in single objective optimization, given the fact that the \emph{tree size} is unbounded, the fitness of a model may improve at the expense of model's complexity. Hence, the experiments were set-up for multiobjective optimization that provides a balance between both objectives as described in Figure~\ref{fig_paretoFront}.    

\section{Results}
\label{sec-res}
Experimental results were classified into three categories: classification, regression, and time-series. Each category has two parts: 1)  First part describes the best and average results obtained from the experiments; 2) Second part describes ensemble results using tabular and graphical form.       

\subsection{Classification dataset}
We chose five classification datasets for evaluating HFNT$^\text{M}$, and the classification accuracy was computed as:
\begin{equation}
\label{eq-accuracy}
f_a = \frac{tp + tn}{tp + fn + fp + tn},
\end{equation}       
where $ tp$ is the total positive samples correctly classified as positive samples, $tn$ is the total negative samples correctly classified as negative samples, $fp$ is the total negative samples incorrectly classified as positive samples, and $fn$ is the total positive samples incorrectly classified as negative samples. Here,  for a binary class classification problem, the positive sample indicates the class labeled with `1' and negative sample indicates class labeled with `0'. Similarly, for a three-class ( $ \omega_1, \omega_2,$  and $\omega_3 $) classification problem, the samples which are labeled as a class $\omega_1$ are set to 1, 0, 0, i.e., set to positive for class $\omega_1$ and negative for $\omega_2, $ and $ \omega_3 $. The samples which are labeled as a class $\omega_2$ are set to 0, 1, 0, and the samples which are labeled as a class $\omega_3$ are set to 0, 0, 1.

\subsubsection{10-Fold CV}
The experiments on classification dataset were conducted in three batches that produced 30 models, and each model was cross-validated using 10-fold CV, in which a dataset is equally divided into 10 sets and the training of a model was repeated 10 times. Each time a distinct set was picked for the testing the models, and the rest of nine set was picked for the training of the model. Accordingly, the obtained results are summarized in Table~\ref{tab_classRes}. Each batch of experiment produced an ensemble system of 10 models whose results are shown in Table~\ref{tab_classEnsembleRes}. 

\begin{figure}[!ht]
	\centering
	\includegraphics[scale = 0.6]{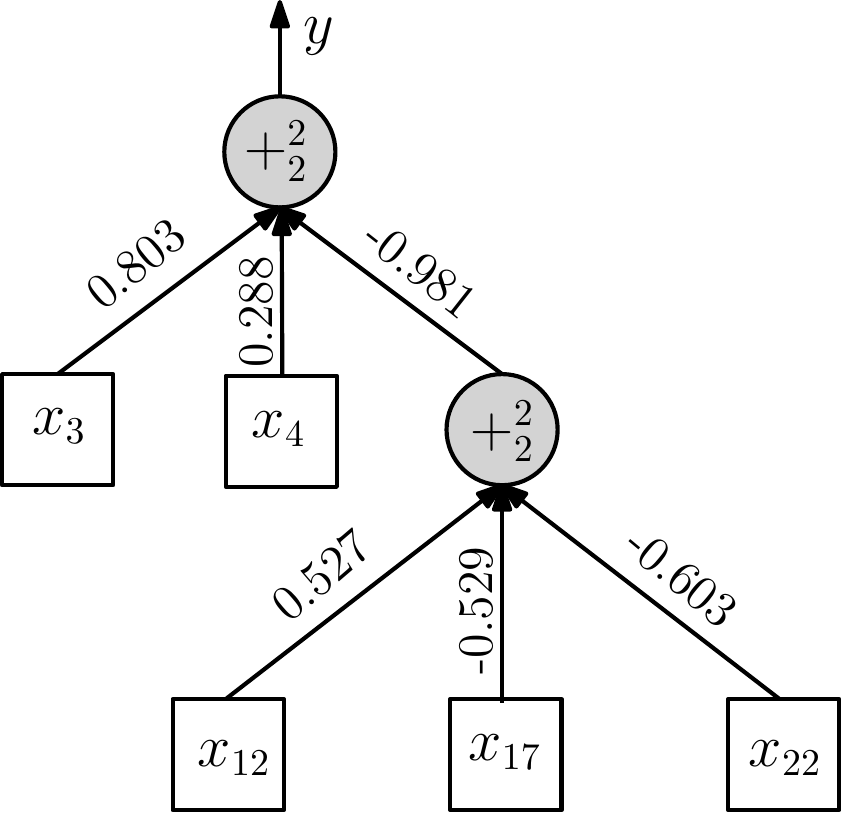}
	\caption{HFNT model of classification dataset WDB (test $ f_a $ = $97.02\%$).}
	\label{fig_fntClasC5}
\end{figure}

The obtained results presented in Table~\ref{tab_classRes} describes the best and mean results of 30 models. We present a comparative study of the best 10-fold CV models results of HFNT$^\text{M}$ and the results reported in the literature in Table~\ref{tab_classRes_comp}. In Table~\ref{tab_classRes_comp}, the results of HDT and FNT~\cite{li2006flexible} were of 10 fold CV results on the test dataset. Whereas, the results of FNT~\cite{chen2006ensemble} was the best test accuracy and not the CV results. The results summarized in Table~\ref{tab_classRes_comp} suggests a comparatively better performance of the proposed HFNT$^\text{M}$ over the previous approaches. For the illustration of a model created by HFNT$^\text{M}$ approach, we chose the best model of dataset WDB that has a test accuracy of $97.02\%$ (shown in Table~\ref{tab_classRes}). A pictorial representation of the WDB model is shown in Figure~\ref{fig_fntClasC5}, where the model's \emph{tree size} is 7, total input features are 5, ($ x_3,x_4, x_{12},x_{17},$ and $ x_{22} $) and the selected activation function is tangent hyperbolic ($ k=2 $) at both the non-leaf nodes. Similarly, we may represent models of all other datasets. 
\begin{table}[!ht]
	\centering
	\footnotesize 
	\caption{Best and mean results of 30 10-fold CV models (300 runs) of HFNT$^\text{M}$}
	\label{tab_classRes}
	\setlength{\tabcolsep}{0.15cm}
	\begin{tabular}{l|ccrr|ccrr}
		\hline
		&  \multicolumn{4}{c|}{Best of 30 models} &  \multicolumn{4}{c}{Mean of 30 models}\\
		\cline{2-9}
		Data & train $ f_a $ & test $ f_a $ & \emph{tree size} & Features & train $ f_a $ & test $ f_a $ & avg. \emph{tree size} & diversity \\
		\hline
		AUS & 87.41\% & 87.39\% & 4  & 3 & 86.59\% & 85.73\% & 5.07 & 0.73 \\
		HRT & 87.41\% & 87.04\% & 8  & 5 & 82.40\% & 80.28\% & 7.50 & 0.70 \\
		ION & 90.92\% & 90.29\% & 5  & 3 & 87.54\% & 86.14\% & 6.70 & 0.83 \\
		PIM & 78.67\% & 78.03\% & 10 & 5 & 71.12\% & 70.30\% & 6.33 & 8.67 \\
		WDB & 97.02\% & 96.96\% & 6  & 5 & 94.51\% & 93.67\% & 7.97 & 0.73 \\
		\hline
	\end{tabular}
\end{table}
             
\begin{table}[!ht]
	\centering
	\footnotesize 
	\caption{Comparative results: 10-fold CV test accuracy $ f_a $ and variance $ \sigma $ of algorithms}
	\label{tab_classRes_comp}
	\setlength{\tabcolsep}{3pt}
	\begin{tabular}{l|rr|rr|rr|rr|rr}
	\hline 
	Algorithms & \multicolumn{2}{c|}{AUS}  & \multicolumn{2}{c|}{HRT}  & \multicolumn{2}{c|}{ION}  & \multicolumn{2}{c|}{PIM}  & \multicolumn{2}{c}{WDB}  \\
	\cline{2-11}
	& test $ f_a $ & $ \sigma  $ & test $ f_a $ & $ \sigma  $ & test $ f_a $ & $ \sigma  $ & test $ f_a $ & $ \sigma  $ & test $ f_a $ & $ \sigma  $   \\
	\hline
	HDT~\cite{li2006flexible}   & 86.96\% & 2.058 & 76.86\% & 2.086 & 89.65\% & 1.624 & 73.95\% & 2.374 &         &      \\
	FNT~\cite{li2006flexible}   & 83.88\% & 4.083 & 83.82\% & 3.934 & 88.03\% & 0.953 & 77.05\% & 2.747 &         &      \\
	FNT~\cite{chen2006ensemble} &  	    &		& 		  &		  & 		&		& 		  &		  & 93.66\% &  n/a \\
	\textbf{HFNT$^\text{M}$} 			& 87.39\% & 0.029 & 87.04\% & 0.053 & 90.29\% & 0.044 & 78.03\% & 0.013 & 96.96\% & 0.005 \\
	\hline
	\end{tabular}
\end{table}

In this work, Friedman test was conducted to examine the significance of the algorithms. For this purpose, the classification accuracy (test results) was considered (Table~\ref{tab_classRes_comp}). The average ranks obtained by each method in the Friedman test is shown in Table~\ref{tab_class_frd_tst}. The Friedman statistic at $ \alpha  = 0.05$ (distributed according to chi-square with 2 degrees of freedom) is 5.991, i.e., $ \chi^2_{(\alpha,2)} = 5.991 $. The obtained test value $ Q $  according to Friedman statistic is 6. Since $ Q > \chi^2_{(\alpha,2)} $, then the \textit{null hypothesis} that ``there is no difference between the algorithms'' is \textit{rejected}. In other words, the computed $ p $-value by Friedman test is  0.049787 which is less than or equal to 0.05, i.e., $ p $-value $ \le \alpha $-value. Hence, we reject the null hypothesis.

Table~\ref{tab_class_frd_tst} describes the significance of differences between the algorithms. To compare the differences between  the best rank algorithm in Friedman test, i.e., between the proposed algorithm HFNT$^\text{M}$ and the other two algorithms, Holm's method~\cite{holm1979simple} was used. Holm's method rejects the hypothesis of equality between the best algorithm (HFNT$^\text{M}$) and other algorithms if the $ p $-value is less than $ \alpha/i $, where $ i $ is the position of an algorithm in a list sorted in ascending order of $ z$-value (Table~\ref{tab_class_post_hoc}). From the post hoc analysis, it was observed that the proposed algorithm HFNT$^\text{M}$ outperformed both HDT~\cite{li2006flexible} and FNT~\cite{li2006flexible} algorithms.

\begin{table}[!htp]
	\centering
	\caption{Average rankings of the algorithms}
	\label{tab_class_frd_tst}
	\begin{tabular}{l l}
		\hline
		Algorithm & Ranking\\
		\hline
		\textbf{HFNT$^\text{M}$} & 1.0\\
		HDT  & 2.5\\
		FNT  & 2.5\\
		\hline
	\end{tabular}
\end{table}

\begin{table}[!htp]
	\centering
	\caption{Post Hoc comparison between HFNT$^\text{M}$ and other algorithms for $\alpha=0.1$}
	\label{tab_class_post_hoc}
	\begin{tabular}{llllll}
		\hline
		$i$ & algorithm & $z$ & $p$ & $ \alpha/i $ & Hypothesis \\
		\hline
		2 & HDT & 2.12132 & 0.033895 & 0.05 & rejected\\
		1 & FNT & 2.12132 & 0.033895 & 0.1 & rejected \\
		\hline
	\end{tabular}
\end{table}

\subsubsection{Ensembles}
The best accuracy and the average accuracy of 30 models presented in Table~\ref{tab_classRes} are the evidence of HFNT$^\text{M}$ efficiency. However, as mentioned earlier, a generalized solution may be obtained by using an ensemble. All 30 models were created in three batches. Hence, three ensemble systems were obtained. The results of those ensemble systems are presented in Table~\ref{tab_classEnsembleRes}, where ensemble results are the accuracies $ f_a $ obtained by weighted majority voting~\eqref{eq_ensmbleWMV}. In Table~\ref{tab_classEnsembleRes}, the classification accuracies $ f_a $ were computed over CV test dataset. From Table~\ref{tab_classEnsembleRes}, it may be observed that high diversity among the ensemble candidates offered comparatively higher accuracy. Hence, an ensemble model may be adopted by examining the performance of an ensemble system, i.e., average \emph{tree size} (complexity) of the candidates within the ensemble and the selected input features. 

An ensemble system created from a genetic evolution and adaptation is crucial for feature selection and analysis. Summarized ensemble results in Table~\ref{tab_classEnsembleRes} gives the following useful information about the HFNT$^\text{M}$ feature selection ability:
1) TSF - total selected features; 2) MSF - most significant (frequently selected) features; and 3) MIF - most infrequently selected features. Table~\ref{tab_classEnsembleRes} illustrates feature selection results.
\begin{table}[!ht]
	\centering
	\footnotesize 
	\caption{Ensemble results (10-fold CV) of each classification dataset}
	\label{tab_classEnsembleRes}
	\setlength{\tabcolsep}{5pt}
	\begin{tabular}{llrrrrcc}	
		\hline 
		Data & Batch & test $ f_a $  & avg. $ D $ & $ div $~\eqref{eq-diversity} & TSF & MSF & MIF \\
		\hline
		AUS & 1 & \textbf{86.96\%} & 5 & 0.7 & 4 & \multirow{3}{2cm}{$x_6$, $x_8$, $x_{10}$, $x_{12}$}  & \multirow{3}{2cm}{$x_1$, $x_2$, $x_3$, $x_{11}$, $x_{14}$}   \\
		& 2 & 85.51\% & 6 & 0.7 & 5 &  &  \\
		& 3 & 86.81\% & 4.2 & 0.8 & 5 &  &   \\
		\hline
		HRT & 1 & 77.41\% & 6.8 & 0.5 & 6 & \multirow{3}{2cm}{$x_3$, $x_4$, $x_{12}$, $x_{13}$} &  \multirow{3}{2cm}{$x_6$} \\
		& 2 & 70.37\% & 7.6 & 0.6 & 9 &  &  \\
		& 3 & \textbf{87.04\%} & 8.1 & 1 & 10 &  &   \\
		\hline
		ION & 1 & 82.86\% & 7.2 & 0.9 & 15 & \multirow{3}{2cm}{$x_2$, $x_4$, $x_5$, $x_{27}$} & \multirow{3}{2cm}{$x_{15}$, $x_{16}$, $x_{18}$, $x_{19}$, $x_{21}$, $x_{23}$, $x_{25}$, $x_{30}$, $x_{32}$}  \\
		& 2 & \textbf{90.29\%} & 7.3 & 1 & 16 &  &   \\
		& 3 & 86.57\% & 5.6 & 0.6 & 6 &  &   \\
		\hline
		PIM & 1 & \textbf{76.32\%} & 6.9 & 1 & 8 & \multirow{3}{2cm}{$x_1$, $x_3$, $x_4$, $x_5$, $x_6$, $x_7$}  & \multirow{3}{2cm}{$x_2$}   \\
		& 2 & 64.74\% & 5.6 & 0.7 & 7 &  &   \\
		& 3 & 64.21\% & 7.4 & 0.9 & 8 &  &   \\
		\hline
		WDB & 1 & 94.29\% & 8.2 & 0.7 & 15 & \multirow{3}{2cm}{$x_{21}$, $x_{22}$, $x_{24}$, $x_{25}$} & \multirow{3}{2cm}{$x_1$, $x_5$, $x_6$, $x_8$, $x_{14}$, $x_{20}$, $x_{30}$}   \\
		& 2 & 93.75\% & 5 & 1 & 15 &  &   \\
		& 3 & \textbf{94.29\%} & 10.7 & 0.5 & 15 &  &   \\		
		\hline
	\end{tabular}	
\end{table}

\subsection{Regression dataset }
\subsubsection{5-Fold CV}
For regression dataset, the performance of HFNT$^\text{M}$ was examined by using 5-fold CV method, in which the dataset was divided into 5 sets, each was 20\% in size, and the process was repeated five times. Each time, four set was used to training and one set for testing. Hence, a total 5 runs were used for each model. As described in~\cite{gacto2014metsk}, MSE $ E = 0.5 \times E$ was used for evaluating HFNT$^\text{M}$, where $ E $ was computed as per~\eqref{eq_mse}. The training MSE is represented as $ E_n $ and test MSE is represented as $ E_t $.  Such setting of MSE computation and cross-validation was taken for comparing the results collected from~\cite{gacto2014metsk}. Table~\ref{tab_regRes} presents results of 5-fold CV of each dataset for 30 models. Hence, each presented result is averaged over a total 150 runs of experiments. Similarly, in Table~\ref{tab_regRes_comp}, a comparison between HFNT$^\text{M}$ and other collected algorithms from literature is shown. It is evident from comparative results that HFNT$^\text{M}$ performs very competitive to other algorithms.  The literature results were averaged over 30 runs of experiments; whereas, HFNT$^\text{M}$ results were averaged of 150 runs of experiments. Hence, a competitive result of HFNT$^\text{M}$ is evidence of its efficiency. 

\begin{figure}[!ht]
	\centering
	\includegraphics[scale = 0.7]{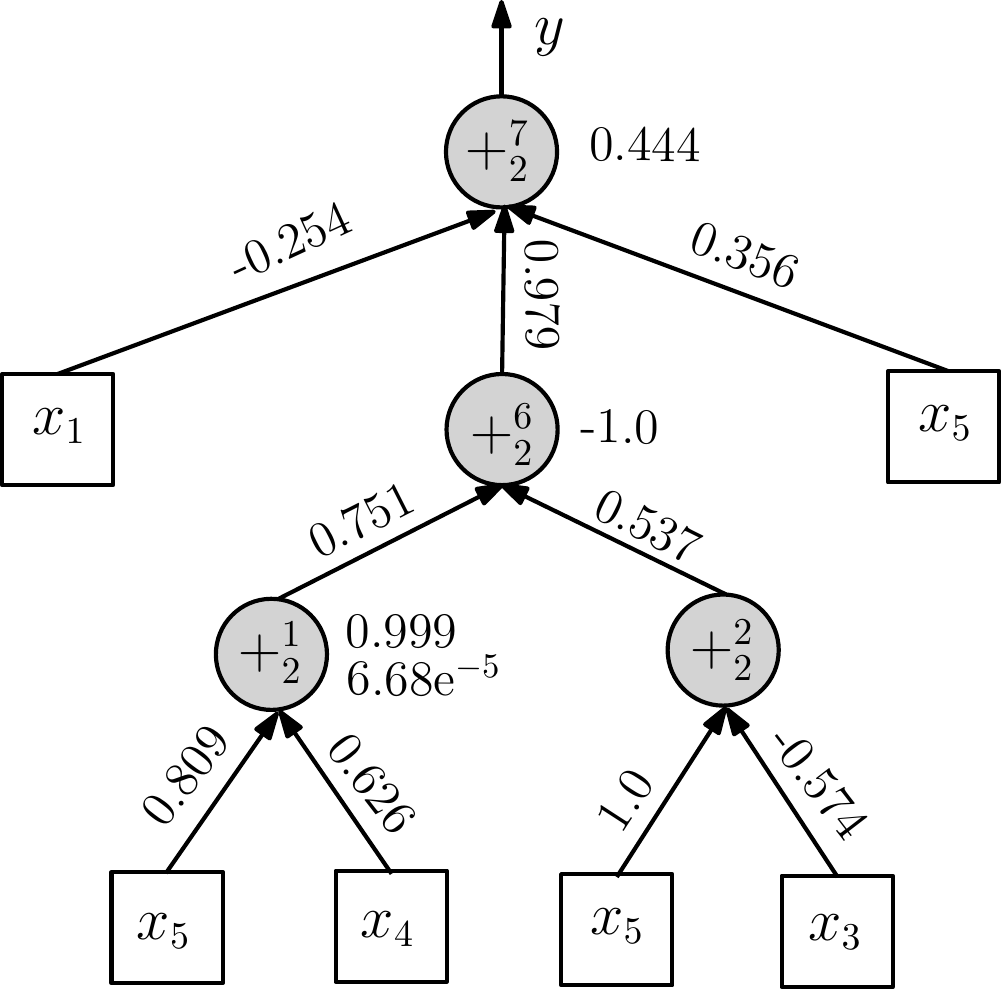}
	\caption{HFNT model of regression dataset  DEE (test MSE $ E_t $ = 0.077).}
	\label{fig_fntRegR4}
\end{figure}

Moreover, HFNT$^\text{M}$ is distinct from the other algorithm mentioned in Table~\ref{tab_regRes_comp} because it performs feature selection and models complexity minimization, simultaneously. On the other hand, the other algorithms used entire available features. Therefore, the result's comparisons were limited to assessing average MSE, where HFNT$^\text{M}$, which gives simple models in comparison to others, stands firmly competitive with the others. An illustration of the best model of regression dataset  DEE is provided in Figure~\ref{fig_fntRegR4}, where the model offered a test MSE $ E_t $ of 0.077, \emph{tree size} equal to 10, and four selected input features ($ x_1$, $x_3$, $ x_4 $, and $x_5$). The selected activation functions were unipolar sigmoid ($ +^7_2 $), bipolar sigmoid ($ +^6_2 $), tangent hyperbolic ($ +^2_2 $), and Gaussian ($ +^1_2 $). Note that while creating HFNT models, the datasets were normalized as described in Table~\ref{tab_fntParameters} and the output of models were denormalized accordingly. Therefore, normalized inputs should be presented to the tree (Figure~\ref{fig_fntRegR4}), and the output $ y $ of the tree (Figure~\ref{fig_fntRegR4}) should be denormalized.
\begin{table}[!ht]
	\centering
	\caption{Best and mean results of 30 5-fold CV models (150 runs) of HFNT$^\text{M}$.}
	\label{tab_regRes}
	\setlength{\tabcolsep}{5pt}
	\begin{tabular}{l|rrrr|rrrr}
		\hline
		&  \multicolumn{4}{c|}{Best of 30 models} &  \multicolumn{4}{c}{Mean of 30 models}\\
		\cline{2-9}
		Data & train $ E_n $ & test $ E_t $ & \emph{tree size} & \#Features & train $ E_n $ & test $ E_t $ & \emph{tree size} & diversity \\
		\hline
		ABL &  2.228 &  2.256 & 14 & 5 & 2.578 & 2.511 & 11.23 & 0.7 \\
		BAS & 198250 & 209582 & 11 & 5 & 261811 & 288688.6 & 7.69 & 0.6 \\
		DEE &  0.076 &  0.077 & 10 & 4 & 0.0807 & 0.086 & 11.7 & 0.7 \\
		ELV$^{*}$  & 8.33 & 8.36 & 11 & 7 & 1.35 & 1.35 & 7.63 & 0.5 \\
		FRD &  2.342 &  2.425 & 6  & 5 & 3.218  &  3.293 & 6.98 & 0.34 \\
		\hline
		\multicolumn{9}{l}{\scriptsize \textbf{Note:} $^{*}$Results of ELV should be multiplied with 10$^{\text{-5}}$}
	\end{tabular}
\end{table}

\begin{table}[!ht]
	\centering
	\footnotesize 
	\caption{Comparative results: 5-fold CV training MSE $ E_n $  and test MSE $ E_t $ of algorithms. }
	\label{tab_regRes_comp}
	\setlength{\tabcolsep}{5pt}
	\begin{tabular}{l|rr|rr|rr|rr|rr}
		\hline 
		Algorithms & \multicolumn{2}{c}{ABL}  & \multicolumn{2}{c}{BAS}  & \multicolumn{2}{c}{DEE} & \multicolumn{2}{c}{ELV$^{*}$} & \multicolumn{2}{c}{FRD}  \\
		\cline{2-11}
		                     &   $E_n$ &  $E_t$ & $E_n$  &   $E_t$ & $E_n$  &   $E_t$ & $E_n$   &  $E_t$ & $E_n$  &   $E_t$ \\
		\hline
		{  MLP}              & -       &  2.694 & -      &  540302          & -      &   0.101          & -       &   2.04  &         &   3.194 \\
		{  ANFIS-SUB}        & 2.008   &  2.733 & 119561 &  1089824         & 3087   &   2083           & 61.417  &   61.35 & 0.085   &   3.158 \\
		{  TSK-IRL}          & 2.581   &  2.642 &        &                  & 0.545  &   882.016        &         &         & 0.433   &   1.419 \\
		{  LINEAR-LMS}       & 2.413   &  2.472 & 224684 &  \textbf{269123} & 0.081  &   0.085          & 4.254   &   4.288 & 3.612   &   3.653 \\
		{  LEL-TSK}          & 2.04    &  2.412 & 9607   &  461402          & 0.662  &   0.682          &         &         & 0.322   &  \textbf{1.07} \\
		{  METSK-HD$^e$}     & 2.205   &  \textbf{2.392} & 47900 & 368820   & 0.03   &   0.103          & 6.75    &  7.02   & 1.075   &  1.887 \\
		{ \textbf{HFNT$^\text{M}$}~\, $^{**}$} & 2.578   &  2.511 & 261811  &  288688.6      & 0.0807 &  \textbf{0.086}  & 1.35  &  \textbf{1.35} & 3.218  &  3.293 \\
		\hline
		\multicolumn{11}{p{14.5cm}}{\scriptsize\textbf{Note: } $^{*}$ELV results should be multiplied with 10$^{\text{-5}}$, $^{**}$HFNT$^\text{M}$ results were averaged over 150 runs compared to MLP, ANFIS-SUB, TSK-IRL, LINEAR-LMS, LEL-TSK, and METSK-HD$^e$, which were averaged over 30 runs.}\\
	\end{tabular}
\end{table}	

For regression datasets, Friedman test was conducted to examine the significance of the algorithms. For this purpose, the best test MSE was considered of the algorithms MLP, ANFIS-SUB, TSK-IRL, LINEAR-LMS, LEL-TSK, and METSK-HD$^e$ from Table~\ref{tab_regRes_comp} and the best test MSE of algorithm HFNT$^\text{M}$ was considered from Table~\ref{tab_regRes}. The average ranks obtained by each method in the Friedman test is shown in Table~\ref{tab_reg_frd_tst}. The Friedman statistic at $ \alpha  = 0.05$ (distributed according to chi-square with 5 degrees of freedom) is 11, i.e., $ \chi^2_{(\alpha,5)} = 11 $. The obtained test value $ Q $  according to Friedman statistic is 11. Since $ Q > \chi^2_{(\alpha,5)} $, then the \textit{null hypothesis} that ``there is no difference between the algorithms'' is \textit{rejected}. In other words, the computed $ p $-value by Friedman test is  0.05 which is less than or equal to 0.05, i.e., $ p $-value $ \le \alpha $-value. Hence, we reject the null hypothesis.
\begin{table}[!htp]
	\centering
	\caption{Average rankings of the algorithms}
	\label{tab_reg_frd_tst}
	\begin{tabular}{l l}
		\hline
		Algorithm & Ranking\\
		\hline
		\textbf{HFNT$^\text{M}$}   & 1.5\\
		METSK-HD$^e$ & 2.75\\
		LEL-TSK      & 3.25\\
		LINEAR-LSM   & 3.5\\
		MLP          & 4.5\\
		ANFIS-SUB    & 5.5\\
		\hline
	\end{tabular}
\end{table}

From the Friedman test, it is clear that the proposed algorithm HFNT$^\text{M}$ performed best among all the other algorithms. However, in the post-hoc analysis presented in Table~\ref{tab_reg_post_hoc} describes the significance of difference between the algorithms. For this purpose, we apply Holm's method~\cite{holm1979simple}, which rejects the hypothesis of equality between the best algorithm (HFNT$^\text{M}$) and other algorithms if the $ p $-value is less than $ \alpha/i $, where $ i $ is the position of an algorithm in a list sorted ascending order of $ z$-value (Table~\ref{tab_reg_post_hoc}). 

In the obtained result, the equality between ANFIS-SUB, MLP and HFNT$^\text{M}$ was rejected, whereas the HFNT$^\text{M}$ equality with other algorithms can not be rejected with $ \alpha = 0.1 $, i.e., with 90\% confidence. However, the $ p $-value shown in Table~\ref{tab_reg_post_hoc} indicates the quality of their performance and the statistical closeness to the algorithm HFNT$^\text{M}$. It can be observed that the algorithm METSK-HD$^e$ performed closer to algorithm HFNT$^\text{M}$, followed by LEL-TSK, and LINEAR-LSM.
\begin{table}[!htp]
	\centering
	\footnotesize
	\caption{Post Hoc comparison between HFNT$^\text{M}$ and other algorithms for $\alpha=0.1$.}
	\label{tab_reg_post_hoc}
	\begin{tabular}{llllll}
		\hline
		$i$ &algorithm & $z$ & $p$ & $ \alpha/i $ & Hypothesis\\
		\hline
		5 &ANFIS-SUB    & 3.023716 & 0.002497 & 0.02 & rejected\\
		4 &MLP          & 2.267787 & 0.023342 & 0.025 & rejected \\
		3 &LINEAR-LSM   & 1.511858 & 0.13057  & 0.033 & \\
		2 &LEL-TSK      & 1.322876 & 0.185877 & 0.05 & \\
		1 &METSK-HD$^e$ & 0.944911 & 0.344704 & 0.1  & \\
		\hline
	\end{tabular}
\end{table}

\subsubsection{Ensembles}
For each dataset, we constructed five ensemble systems by using 10 models in each batch. In each batch, 10 models were created and cross-validated using $ 5\times2 $-fold CV. In $ 5\times2 $-fold CV, a dataset is randomly divided into two equal sets: A and B. Such partition of the dataset was repeated five times and each time when the set A was presented for training, the set B was presented for testing, and vice versa. Hence, total 10 runs of experiments for each model was performed. The collected ensemble results are presented in Table~\ref{tab_regEnsembleRes}, where ensemble outputs were obtained by using weighted arithmetic mean as mentioned in~\eqref{eq_ensmbleWAM}.

\begin{figure}
	\centering
	\subfigure[Dataset ABL. $ r_t = 0.75$]
	{
		\includegraphics[width = 0.45\textwidth]{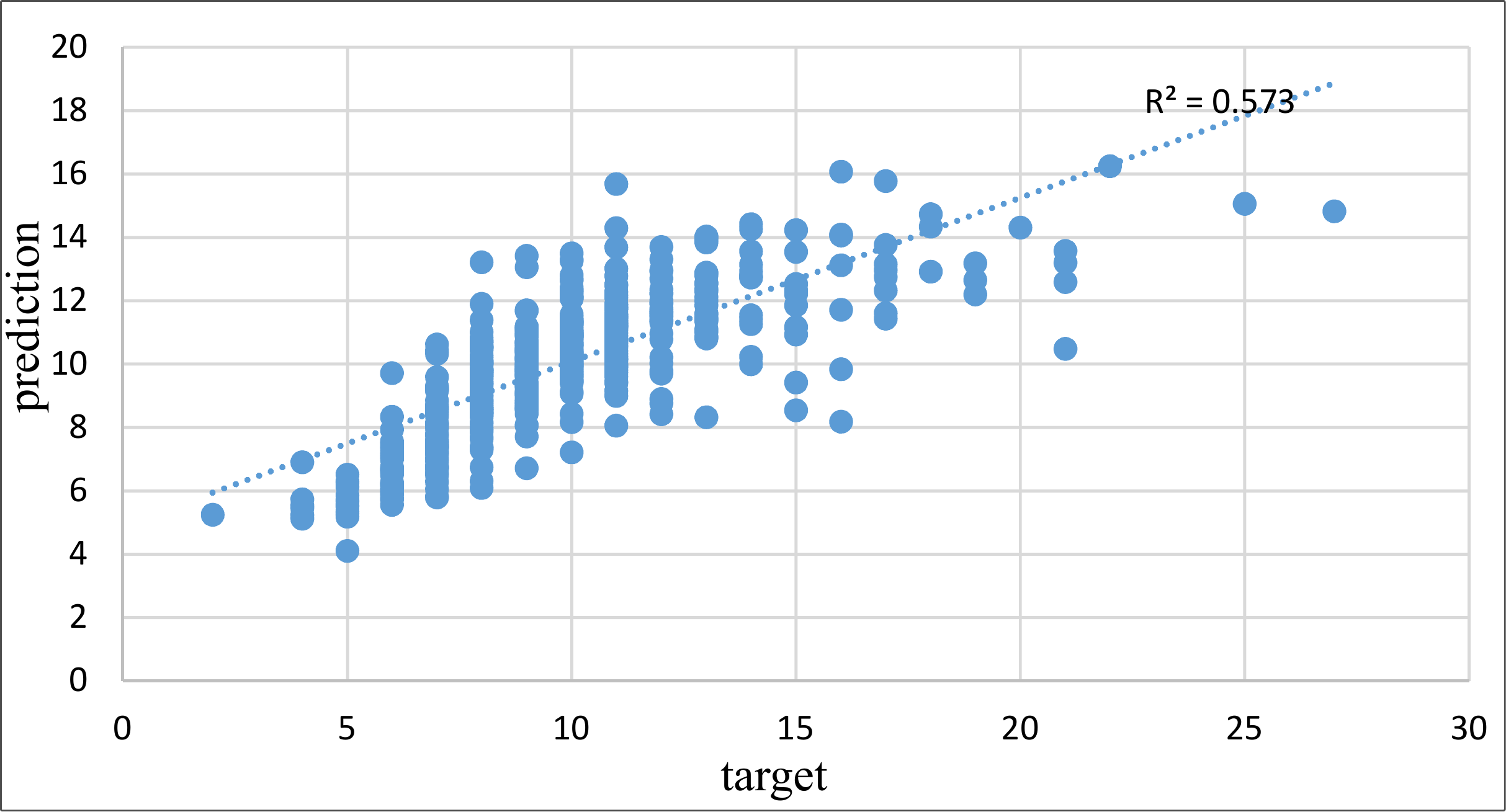}
		\label{fig_reg_1}%
	}
	\subfigure[Dataset BAS. $ E_t = 0.81$]
	{
		\includegraphics[width = 0.45\textwidth]{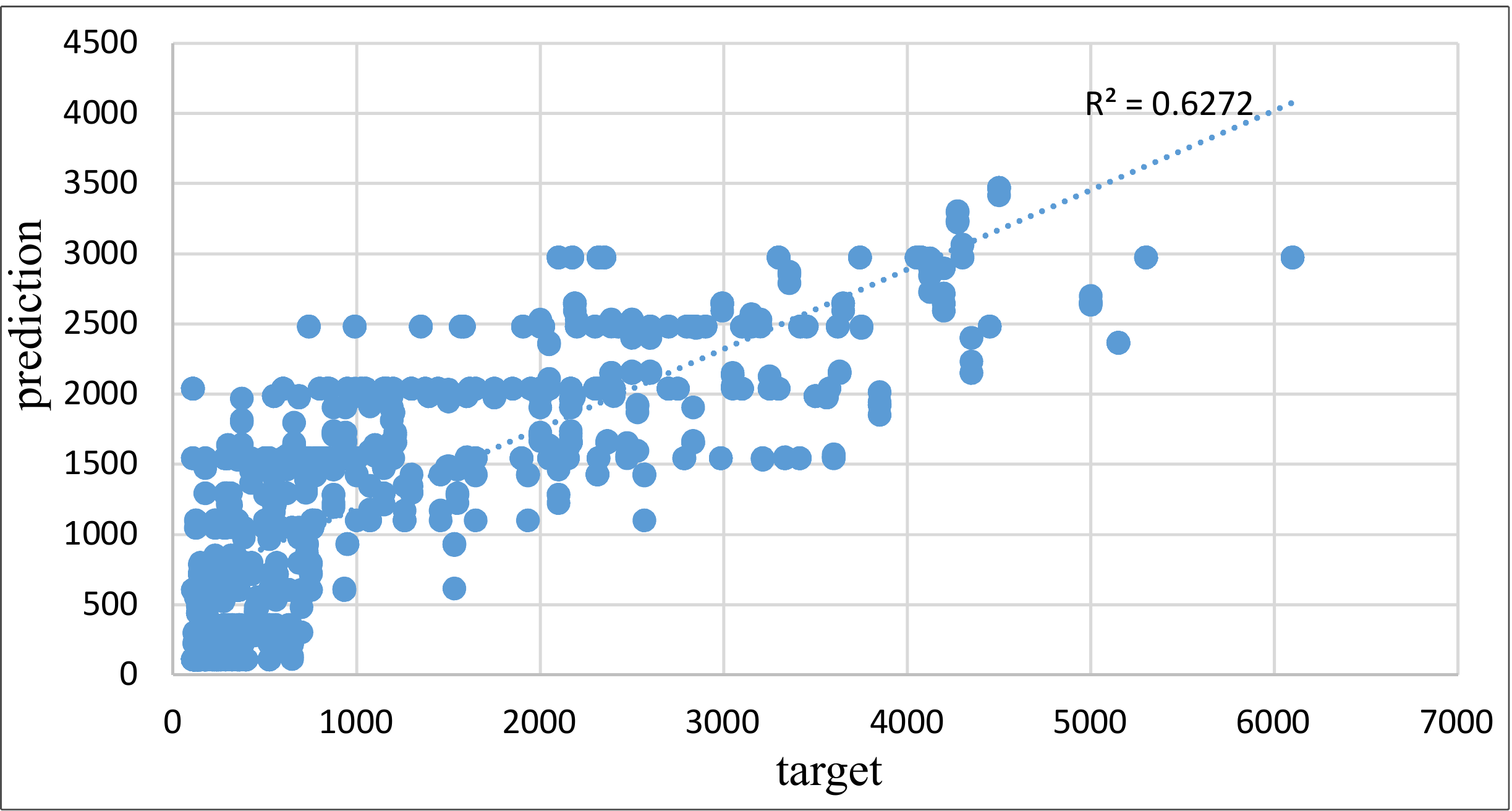}
		\label{fig_reg_2}%
	}
	\subfigure[Dataset DEE. $ r_t = 0.88$]
	{
		\includegraphics[width = 0.45\textwidth]{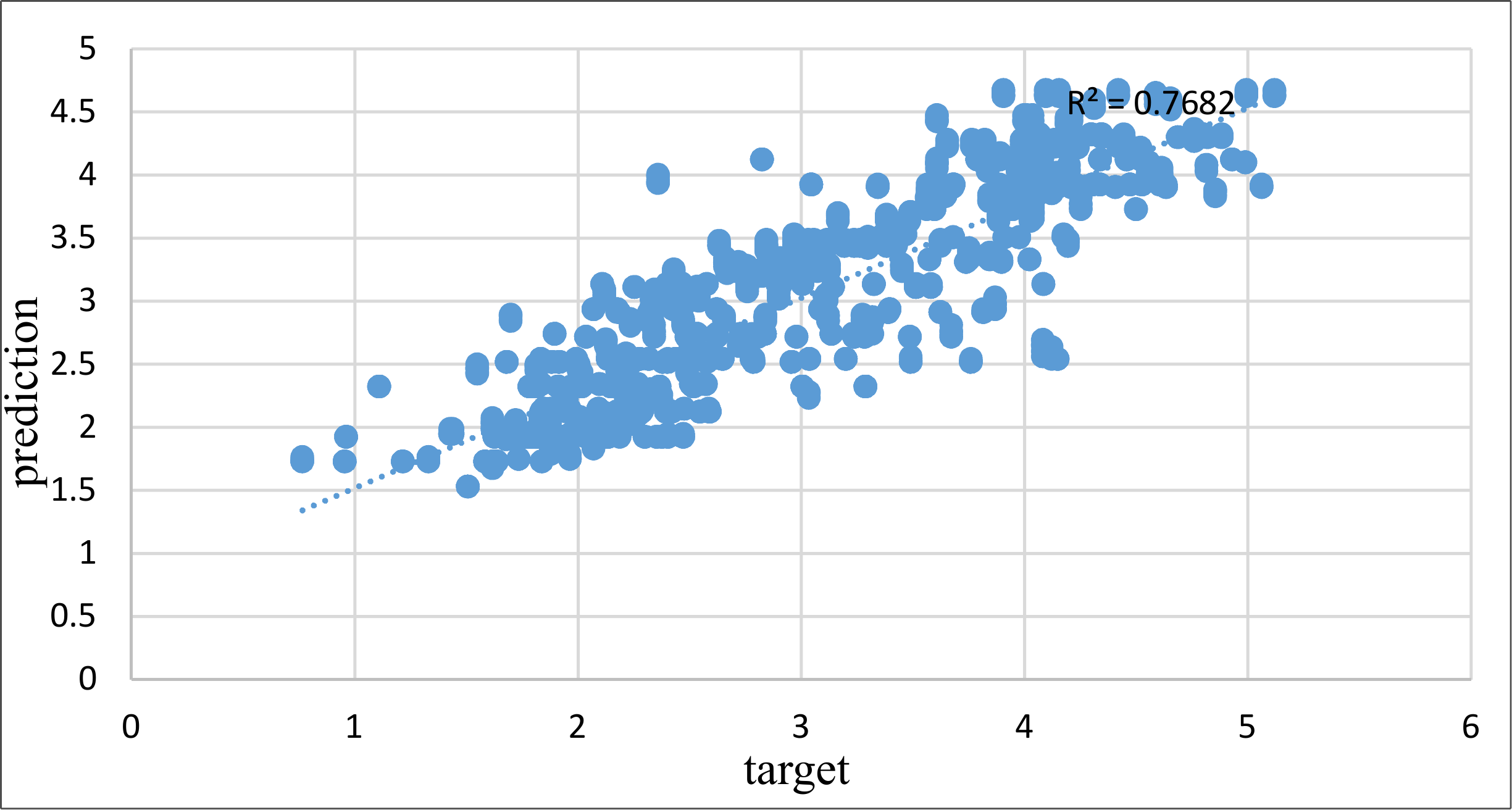}
		\label{fig_reg_3}%
	}
	\subfigure[Dataset EVL. $ r_t = 0.72$]
	{
		\includegraphics[width = 0.45\textwidth]{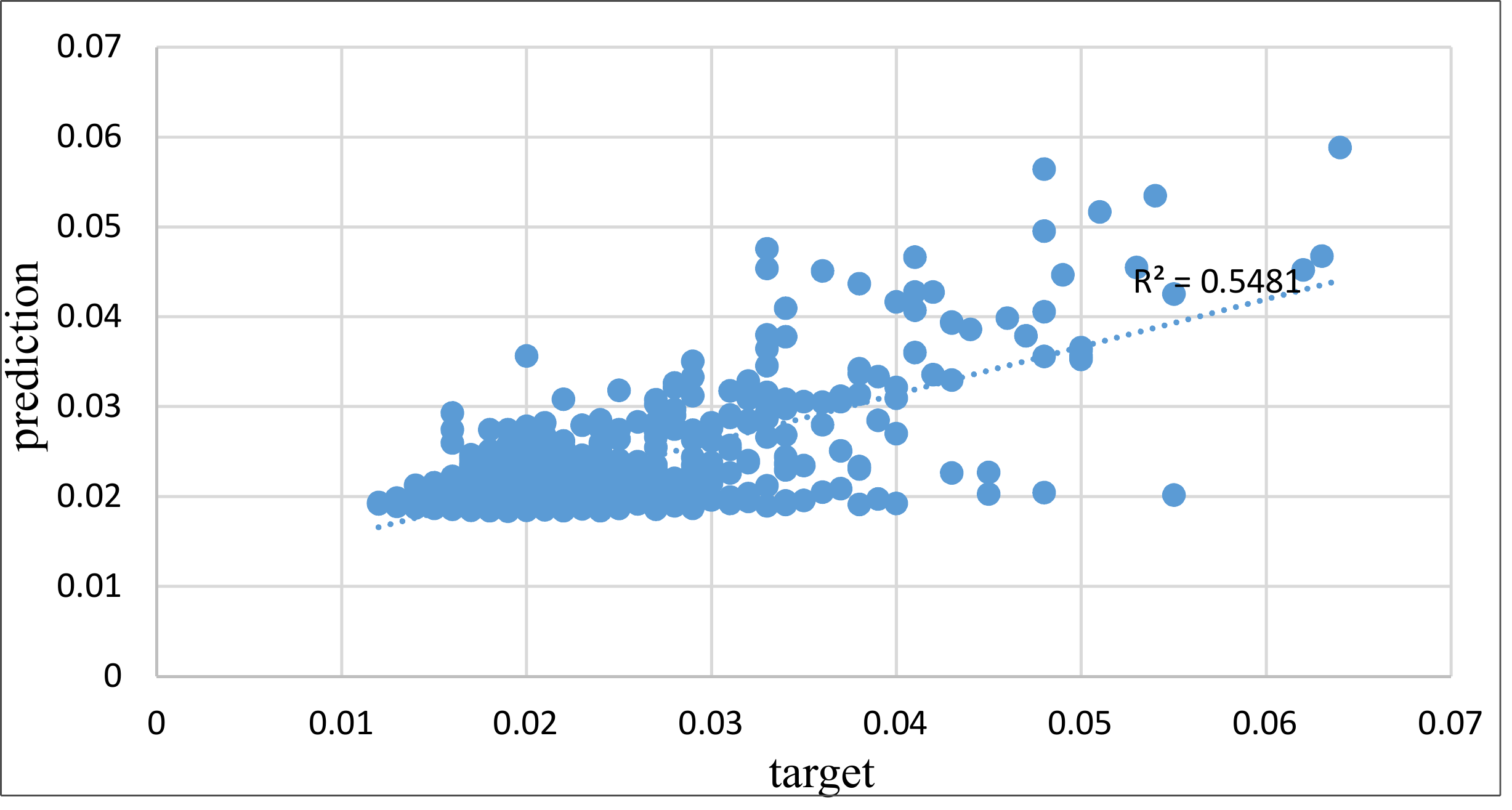}
		\label{fig_reg_4}%
	}
	\subfigure[Dataset FRD. $ r_t = 0.91$]
	{
		\includegraphics[width = 0.45\textwidth]{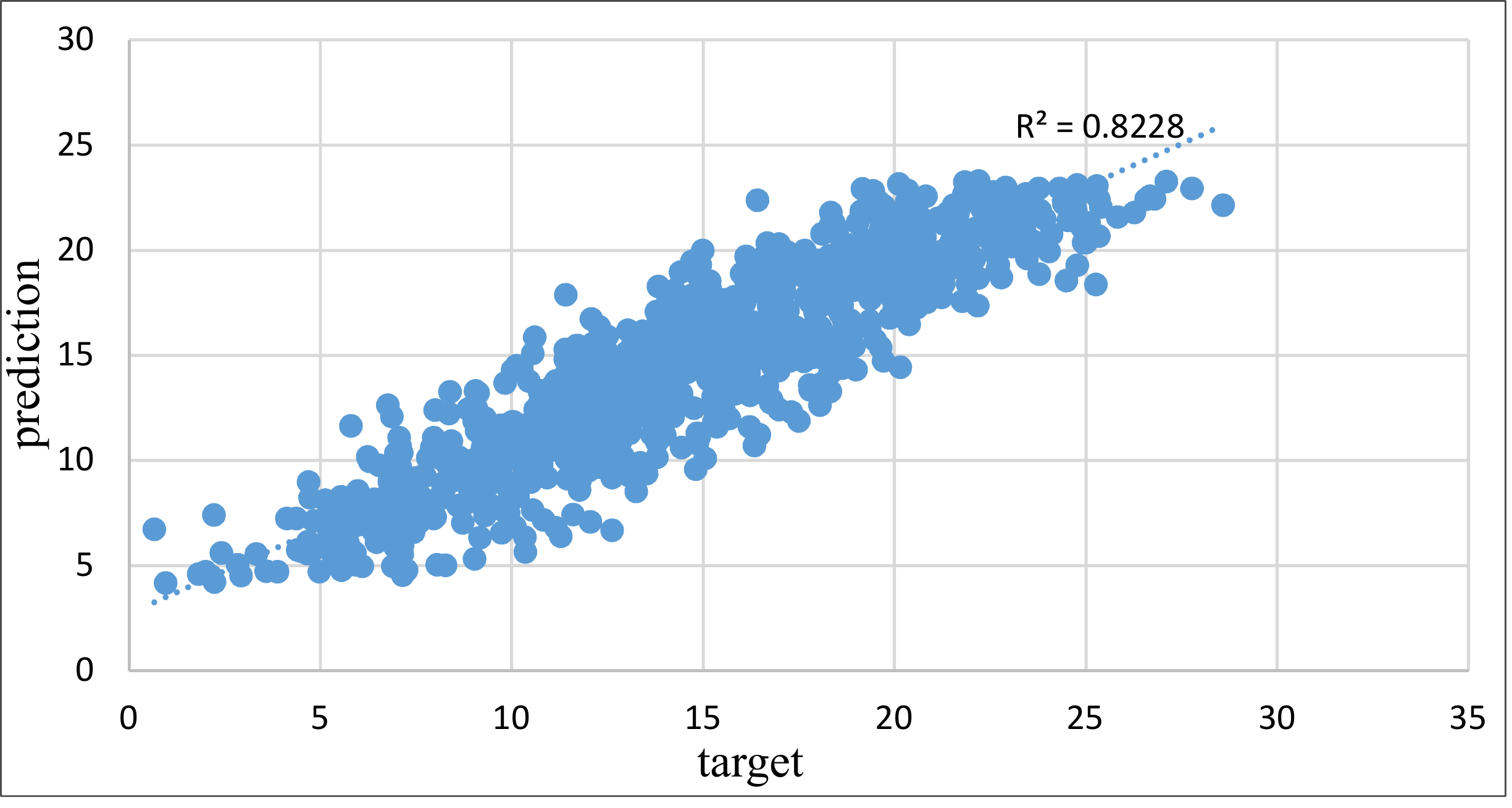}
		\label{fig_reg_5}%
	}
	\caption{Regression plots of the best ensemble batches on datasets R1, R2, R3, R4, and R5.}
	\label{fig_regPlots}
\end{figure}

The weights of models were computed by using DE algorithm, where the parameter setting was similar to the one mentioned in classification dataset. Ensemble results shown in Table~\ref{tab_regEnsembleRes} are MSE and correlation coefficient computed on CV test dataset. From ensemble results, it can be said that the ensemble with higher diversity offered better results than the ensemble with lower diversity. The models of the ensemble were examined to evaluate MSF and MIF presented in Table~\ref{tab_regEnsembleRes}. A graphical illustration of ensemble results is shown in Figure~\ref{fig_regPlots} using scattered (regression) plots, where a scatter plots show how much one variable is affected by another (in this case  model's and desired outputs). Moreover, it tells the relationship between two variables, i.e., their correlation. Plots shown in Figure~\ref{fig_regPlots} represents the best ensemble batch (numbers indicated bold in Table~\ref{tab_regEnsembleRes}) four, five, three, four and five where MSEs are 2.2938, 270706, 0.1085, 1.10$E-$05 and 2.3956, respectively. The values of $ r^2 $ in plots tell about the regression curve fitting over CV test datasets. In other words, it can be said that the ensemble models were obtained with generalization ability.   
\begin{table}
	\centering
	\scriptsize 
	\caption{Ensemble test MSE $ E_t $ computed for $ 5 \times 2 $-fold CV of 10 model in each batch}
	\label{tab_regEnsembleRes}
	\setlength{\tabcolsep}{3pt}
	\begin{tabular}{lclrrrrll}	
		\hline 
		Data & batch & MSE $ E_t $ & $ r_t $ & avg.~$ D $ & $ div $~\eqref{eq-diversity} & TSF & MSF & MIF \\ 
		\hline ABL 
		& 1 & 3.004 & 0.65 & 5 & 0.1 & 3 & \multirow{5}{1.5cm}{$x_2$, $x_3$, $x_5$, $x_6$} &  \multirow{5}{1.5cm}{$x_1$} \\
		& 2 & 2.537 & 0.72 & 8.3 & 1 & 7 &  &  \\
		& 3 & 3.042 & 0.65 & 8.5 & 0.5 & 5 &  &  \\
		& \textbf{4} & {2.294} & \textbf{0.75} & 10.7 & 1 & 7 &  &  \\
		& 5 & 2.412 & 0.73 & 11.2 & 0.7 & 7 &  &  \\
		\hline BAS$^{*}$ 
		& 1 & 2.932 & 0.79 & 5.6 & 0.3 & 5 & \multirow{5}{1.5cm}{$x_3$, $x_7$, $x_8$, $x_9$, $x_{11}$, $x_{13}$} & \multirow{5}{1.5cm}{$x_1$, $x_2$, $x_5$, $x_6$, $x_{10}$} \\
		& 2 & 3.275 & 0.76 & 8.2 & 0.3 & 6 &  &  \\
		& 3 & 3.178 & 0.77 & 5 & 0.2 & 7 &  &  \\
		& 4 & 3.051 & 0.78 & 5.7 & 0.3 & 5 &  &  \\
		& \textbf{5} & {2.707} & \textbf{0.81} & 7.3 & 0.7 & 9 &  &  \\
		\hline DEE 
		& 1 & 0.112 & 0.88 & 4.3 & 0.2 & 4 & \multirow{5}{1.5cm}{$x_1$, $x_3$, $x_4$, $x_5$, $x_6$} & \multirow{5}{1.5cm}{$x_2$} \\
		& 2 & 0.115 & 0.88 & 8.9 & 0.6 & 6 &  &  \\
		& \textbf{3} & {0.108} & \textbf{0.88} & 5.4 & 0.5 & 3 &  &  \\
		& 4 & 0.123 & 0.87 & 10.8 & 0.9 & 5 &  &  \\
		& 5 & 0.111 & 0.88 & 5.2 & 0.6 & 4 &  &  \\
		\hline EVL$^{**}$ 
		& 1 & 1.126 & 0.71 & 9.3 & 0.1 & 12 & \multirow{5}{1.5cm}{$x_1$, $x_3$, $x_4$, $x_6$, $x_{17}$} & \multirow{5}{1.5cm}{$x_7$, $x_8$, $x_{12}$} \\
		& 2 & 1.265 & 0.67 & 9.6 & 0.1 & 12 &  &  \\
		& 3 & 1.124 & 0.71 & 10.4 & 0.1 & 15 &  &  \\
		& \textbf{4} & {1.097} & \textbf{0.72} & 9.2 & 0.2 & 10 &  &  \\
		& 5 & 2.047 & 0.31 & 3.8 & 0.4 & 3 &  &  \\
		\hline FRD 
		& 1 & 3.987 & 0.86 & 6.2 & 0.2 & 4 & \multirow{5}{1.5cm}{$x_1$, $x_2$, $x_4$, $x_5$} & \multirow{5}{1.5cm}{$x_3$} \\
		& 2 & 4.154 & 0.83 & 8 & 0.2 & 4 &  &  \\
		& 3 & 4.306 & 0.83 & 5.2 & 0.4 & 5 &  &  \\			
		& 4 & 3.809 & 0.86 & 7.8 & 0.5 & 4 &  &  \\
		& \textbf{5} & {2.395} & \textbf{0.91} & 7.7 & 0.4 & 5 &  &  \\		
		\hline
		\multicolumn{9}{l}{\tiny\textbf{Note: } $^{*}$BAS results should be multiplied with 10$^{\text{5}}$, $^{**}$ELV results should be multiplied with 10$^{\text{-5}}$.}\\
	\end{tabular}	
\end{table}	

\subsection{Time-series dataset}
\subsubsection{2-Fold CV}
In literature survey, it was found that efficiency of most of the FNT-based models was evaluated over time-series dataset. Mostly, Macky-Glass (MGS) dataset was used for this purpose. However, only the best-obtained results were reported. For time-series prediction problems, the performances were computed using the root of mean squared error (RMSE), i.e., we took the square root of $ E $ given in~\eqref{eq_mse}. Additionally, correlation coefficient~\eqref{eq_corr} was also used for evaluating algorithms performance. 

For the experiments, first 50\% of the dataset was taken for training and the rest of 50\% was used for testing. Table~\ref{tab_tsRes} describes the results obtained by HFNT$^\text{M}$, where $ E_n $ is RMSE for training set and $ E_t $ is RMSE for test-set. The best test RMSE obtained by HFNT$^\text{M}$ was $ E_t = 0.00859$ and $ E_t = 0.06349$ on datasets MGS and WWR, respectively. HFNT$^\text{M}$ results are competitive with most of the algorithms listed in Table~\ref{tab_tsRes_comp}. Only a few algorithms such as LNF and FWNN-M reported better results than the one obtained by HFNT$^\text{M}$. FNT based algorithms such as FNT~\cite{chen2005time} and FBBFNT-EGP\&PSO reported RMSEs close to the results obtained by HFNT$^\text{M}$. The average RMSEs and its variance over test-set of 70 models were 0.10568 and 0.00283, and 0.097783 and 0.00015 on dataset MGS and WWR, respectively. The low variance indicates that most models were able to produce results around the average RMSE value. The results reported by other function approximation algorithms (Table~\ref{tab_tsRes}) were merely the best RMSEs. Hence, the robustness of other reported algorithm cannot be compared with the HFNT$^\text{M}$. However, the advantage of using HFNT$^\text{M}$ over other algorithms is evident from the fact that the average complexity of the predictive models were 8.15 and 8.05 for datasets MGA and WWR, respectively. 

\begin{figure}[!h]
	\centering
	\includegraphics[scale = 0.5]{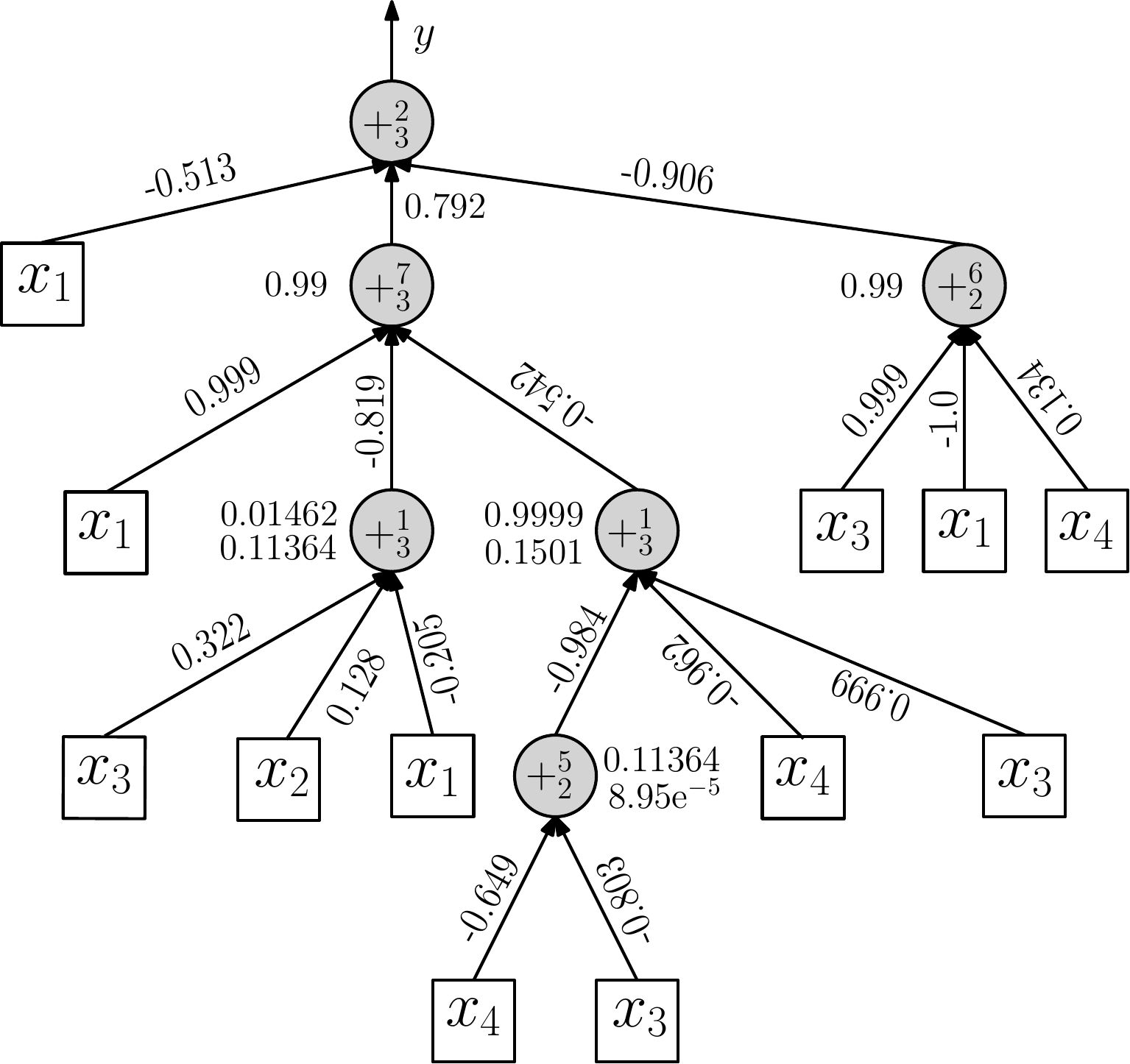}
	\caption{HFNT model of time-series dataset WWR (RMSE = 0.063489).}
	\label{fig-fntTsT1}
\end{figure}

The best model obtained for dataset WWR is shown in Figure~\ref{fig-fntTsT1}, where the \emph{tree size} is equal to 17 and followings are the selected activation functions: tangent hyperbolic, Gaussian, unipolar sigmoid, bipolar sigmoid and linear tangent hyperbolic. The selected input features in the tree (Figure~\ref{fig-fntTsT1}) are $x_1$, $x_2$, $x_3$ and $x_4$. Since in time series category experiment, we have only two datasets and for each dataset HFNT$ ^\text{M} $ was compared with different models from literature. Hence, the statistical test was not conducted in this category because differences between algorithms are easy to determine from  Table~\ref{tab_tsRes_comp}. 
\begin{table}[!h]
	\centering
	\footnotesize 
	\caption{Best and mean results 2-fold CV training RMSE $ E_n $ and test RMSE $ E_t $. }
	\label{tab_tsRes}
	\setlength{\tabcolsep}{5pt}
	\begin{tabular}{lrrrlrrr}	
		\hline
		&  \multicolumn{4}{c}{Best of 70 models} &  \multicolumn{3}{c}{Mean of 70 models}\\
		\cline{2-8}
		Data &   $ E_n $ &   $ E_t $ & $ D $ & Features  &   $ E_n $ &   $ E_t $ & $ D $ \\
		\hline
		MGS &  0.00859  & 0.00798 & 21 & 4 & 0.10385  & 0.10568 & 8.15 \\
		WWR &  0.06437  & 0.06349 & 17 & 4 & 0.10246 &  0.09778 & 8.05 \\ 
		\hline				
	\end{tabular}	
\end{table}

\begin{table}[!h]
	\centering
	\footnotesize 
	\caption{Comparative results: training RMSE $ E_n $ and test RMSE $ E_t $ for 2-fold CV.}
	\label{tab_tsRes_comp}
	\setlength{\tabcolsep}{5pt}
	\begin{tabular}{l|rr|rr}	
		\hline 
		Algorithms & \multicolumn{2}{c|}{MGS} & \multicolumn{2}{c}{WWR} \\
		\cline{2-5}
		&	 $ E_n $ &  $ E_t $ &  $  E_n $ &  $ E_t $ \\														
		\hline				       
		CPSO     		       	& 0.0199  & 0.0322  &   &  \\ 
		PSO-BBFN 			   	& -       & 0.027   &   &  \\
		HCMSPSO 				& 0.0095  & 0.0208  &   &  \\
		HMDDE-BBFNN            	& 0.0094  & 0.017   &   &  \\
		G-BBFNN                	& -       & 0.013   &   &  \\
		Classical RBF		    & 0.0096  & 0.0114  &   &  \\
		FNT~\cite{chen2005time} & 0.0071  & 0.0069  &   &  \\
		FBBFNT-EGP\&PSO        	& 0.0053  & 0.0054  &   &  \\
		FWNN-M 					& 0.0013  & 0.00114 &   &  \\
		LNF 		                & 0.0007  & 0.00079 &   &  \\
		BPNN 	     	 	   	& 	-	  &	-		&  -   &  0.200  \\
		EFuNNs   		  	   	& 	-	  &	-    	&  0.1063  &  0.0824 \\
		\textbf{HFNT$^\text{M}$ } 		& 0.00859  & 0.00798 &  0.064377 & \textbf{0.063489} \\
		\hline				
	\end{tabular}	
\end{table}
\subsubsection{Ensembles}
\begin{figure}[!h]
	\centering
	\subfigure[Dataset MGS $ E_t = 01815$]
	{
		\includegraphics[width = 0.6\textwidth]{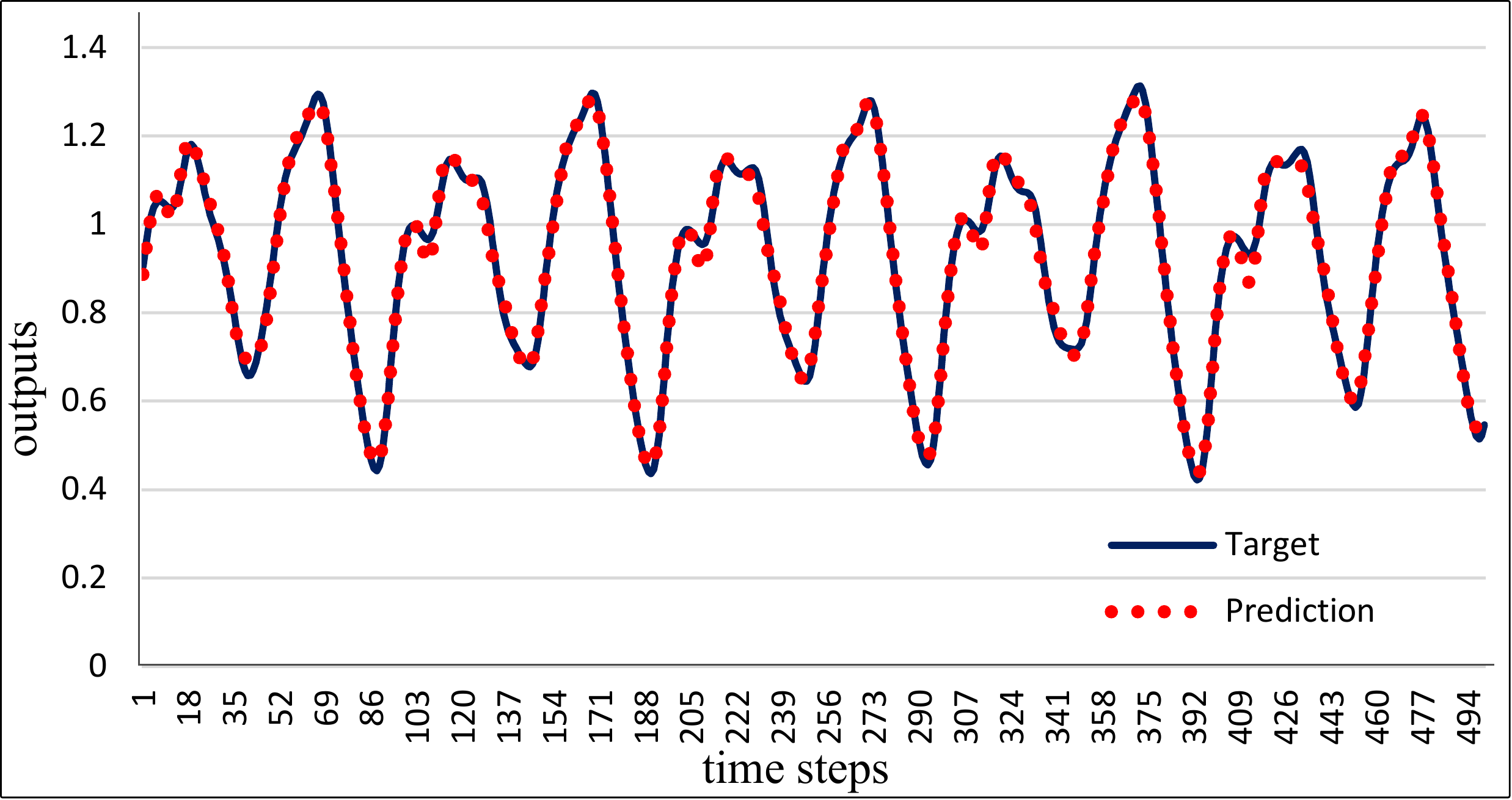} 
		\label{fig_ts_1}%
	}
	\subfigure[Dataset WWR $ E_t = 0.06328$]
	{
		\includegraphics[width = 0.6\textwidth]{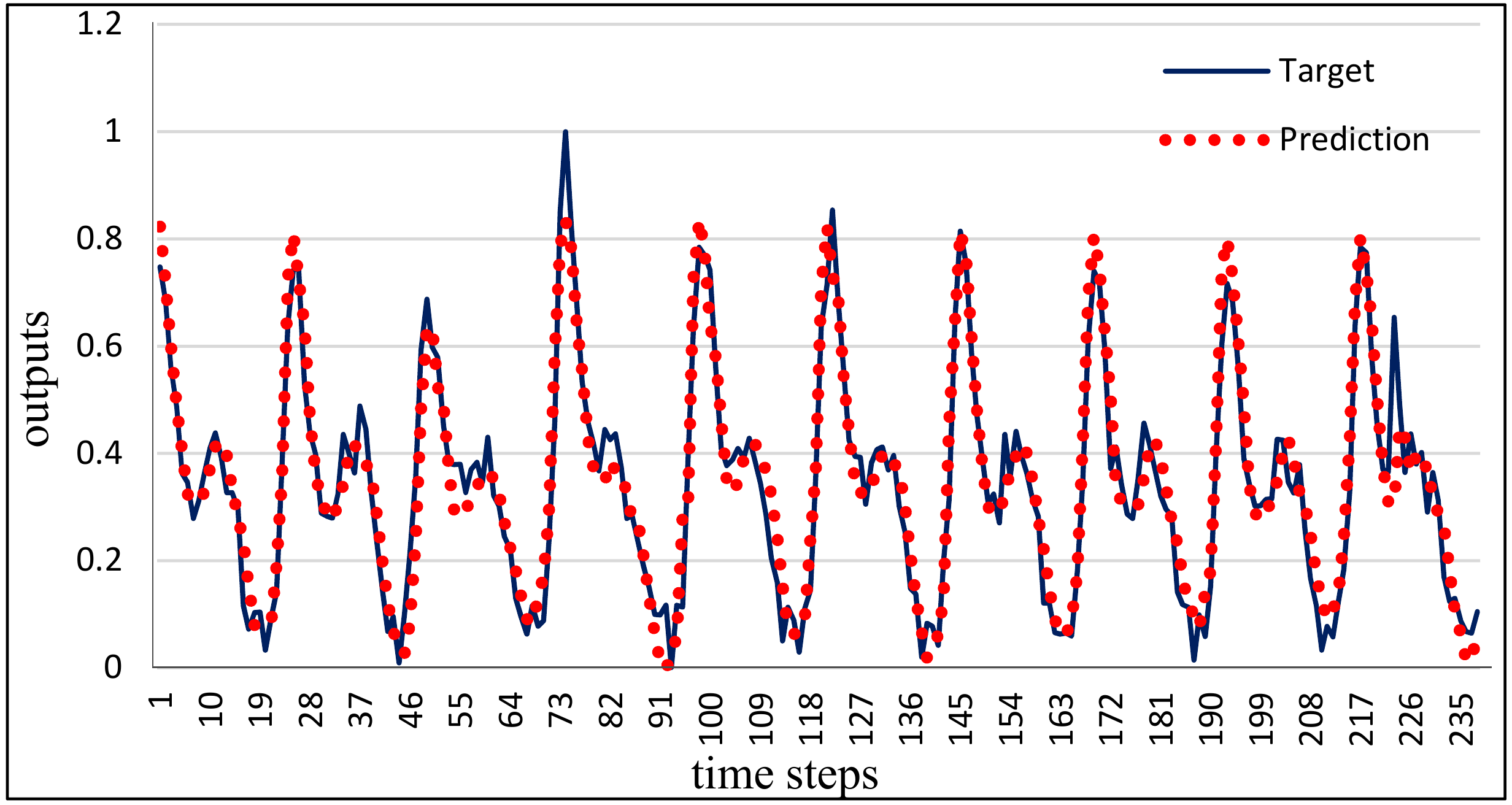}
		\label{fig_ts_2}%
	}
	\caption{Target versus  prediction plot obtained for time-series datasets MGS and WWR.}
	\label{fig_tsPlots}
\end{figure}

The ensemble results of time-series datasets are presented in Table~\ref{tab_tsEnsembleRes}, where the best ensemble system of dataset MGS (marked bold in Table~\ref{tab_tsEnsembleRes}) offered a test RMSE $ E_t = 0.018151$ with a test correlation coefficient $ r_t = 0.99$. Similarly, the best ensemble system of dataset WWR (marked bold in Table~\ref{tab_tsEnsembleRes}) offered a test RMSE $ E_t = 0.063286$  with a test correlation coefficient $ r_t =  0.953$. However, apart from the best results, most of the ensemble produced low RMSEs, i.e., high correlation coefficients. The best ensemble batches (marked bold in Table~\ref{tab_tsEnsembleRes}) of dataset MGS and WWR were used for graphical plots in Figure~\ref{fig_tsPlots}. A one-to-one fitting of target and prediction values is the evidence of a high correlation between model's output and desired output, which is a significant indicator of model's efficient performance.     
   
\begin{table}[!h]
	\centering
	\footnotesize 
	\caption{Ensemble results computed for 50\% test samples of time-series datasets}
	\label{tab_tsEnsembleRes}
	\setlength{\tabcolsep}{5pt}
	\begin{tabular}{lclrrrrrr}	
		\hline 
		Data & batch & $ E_t $ & $ r_t $ & avg. \emph{tree size} & $ div $~\eqref{eq-diversity} & TSF & MSF & MIF \\
		\hline MGS 
		& \textbf{1} & {0.018} & \textbf{0.99} & 9.4 & 0.6 & 4 & $x_1$, $x_3$, $x_4$ & - \\
		& 2 & 0.045 & 0.98 & 5.8 & 0.2 & 3 &  &  \\
		& 3 & 0.026 & 0.99 & 15.2 & 0.5 & 3 &  &  \\
		& 4 & 0.109 & 0.92 & 5.1 & 0.4 & 3 &  &  \\
		& 5 & 0.156 & 0.89 & 7 & 0.2 & 3 &  &  \\
		& 6 & 0.059 & 0.97 & 8.2 & 0.5 & 3 &  &  \\
		& 7 & 0.054 & 0.98 & 6.4 & 0.4 & 4 &  &  \\
		\hline WWR 
		& 1 & 0.073 & 0.94 & 5 & 0.1 & 3 & $x_1$, $x_2$ & - \\
		& 2 & 0.112 & 0.85 & 6 & 0.2 & 2 &  &  \\
		& 3 & 0.097 & 0.91 & 10.6 & 0.3 & 4 &  &  \\
		& 4 & 0.113 & 0.84 & 5 & 0.1 & 2 &  &  \\
		& \textbf{5} & {0.063} & \textbf{0.96} & 14.4 & 0.9 & 4 &  &  \\
		& 6 & 0.099 & 0.89 & 8.5 & 0.7 & 3 &  &  \\
		& 7 & 0.101 & 0.88 & 6.9 & 0.4 & 3 &  &  \\
		\hline
		\multicolumn{9}{p{10cm}}{{\scriptsize \textbf{Note:} $ E_t $, $ r_t $, and $ div $ indicate test RMSE, test correlation coefficient, and diversity, respectively}}				
	\end{tabular}	
\end{table}	

\section{Discussions}
\label{sec-dis}
HFNT$^\text{M}$ was examined over three categories of datasets: classification, regression, and time-series. The results presented in Section~\ref{sec-res}, clearly suggests a superior performance of HFNT$^\text{M}$ approach. In HFNT$^\text{M}$ approach, MOGP guided an initial HFNT population towards Pareto-optimal solutions, where HFNT final population was a mixture of heterogeneous HFNTs. Alongside, accuracy and simplicity, a Pareto-based multiobjective approach ensured diversity among the candidates in final population. Hence, HFNTs in the final population were fairly accurate, simple, and diverse. Moreover, HFNTs in the final population were diverse according to structure, parameters, activation function, and input feature. Hence, the model's selection from Pareto-fronts, as indicated in Section~\ref{sec-exp}, led to a good ensemble system.  

\begin{table}[!h]
	\centering
	\footnotesize 
	\caption{Performance of activation functions during the best performing ensembles}
	\label{tab_act_fun_perfm}
	\setlength{\tabcolsep}{5pt}
	\begin{tabular}{lccccccc}	
		\hline
		& \multicolumn{7}{c}{activation function ($k$)} \\
		\cline{2-8}
		Data & 1 & 2 & 3 & 4 & 5 & 6 & 7 \\
		\hline
		AUS & 10 & - & - & 2 & - & - & - \\
		HRT & 10 & - & 9 & 4 & - & 5 & 3 \\
		ION & 6 & 5 & - & - & 2 & 4 & 4 \\
		PIM & 3 & 8 & 2 & 5 & 2 & 1 & - \\
		WDB & - & 3 & - & 7 & 8 & 10 & 8 \\
		ABL & 2 & 10 & - & - & - & 10 & - \\
		BAS & 2 & 5 & - & - & 2 & 10 & - \\
		DEE & - & 6 & 6 & 4 & 4 & 10 & - \\
		EVL & 10 & 5 & - & 3 & - & - & 6 \\
		FRD & 10 & 10 & - & - & - & - & - \\
		MGS & 4 & 1 & - & 2 & 1 & 10 & 10 \\
		WWR & 10 & - & 4 & - & 4 & 7 & - \\
		\hline
		Total & 67 & 53 & 21 & 27 & 23 & 67 & 31 \\
		\hline	
		\multicolumn{8}{l}{{\scriptsize \textbf{Note:} 67 is the best and  21 is the worst}}			
	\end{tabular}	
\end{table}	
HFNT$^\text{M}$ was applied to solve classification, regression, and time-series problems. Since HFNT$^\text{M}$ is stochastic in nature, its performance was affected by several factors: random generator algorithm, random seed, the efficiency of the meta-heuristic algorithm used in \emph{parameter-tuning} phase, the activation function selected at the nodes, etc. Therefore, to examine the performance of HFNT$^\text{M}$, several HFNT-models were created using different random seeds and the best and average \emph{approximation error} of all created models were examined. In Section~\ref{sec-res}, as far as the best model is concerned, the performance of HFNT$^\text{M}$ surpass other approximation models mentioned from literature. Additionally, in the case of each dataset, a very low average value (high accuracy in the case of classification and low approximation errors in case of regression and time-series) were obtained, which significantly suggests that HFNT$^\text{M}$ often led to good solutions. Similarly, in the case of the ensembles, it is clear from the result that combined output of diverse and accurate candidates offered high quality (in terms of generalization ability and accuracy) approximation/prediction model. From the results, it is clear that the final population of HFNT$^\text{M}$ offered the best ensemble when the models were carefully examined based on \emph{approximation error},  average complexity (\emph{tree size}), and selected features. 

Moreover, the performances of the best performing activation functions were examined. For this purpose, the best ensemble system obtained for each dataset were considered. Accordingly, the performance of activation functions was evaluated as follows. The best ensemble system of each dataset had 10 models; therefore, in how many models (among 10) an activation function $ k $ appeared, was counted. Hence, for a dataset, if an activation function appeared in all models of an ensemble system, then the total count was 10. Subsequently, counting was performed for all the activation functions for the best ensemble systems of all the datasets. Table~\ref{tab_act_fun_perfm}, shows the performance of the activation functions. It can be observed that the activation function Gaussian ($k=1$) and Bipolar Sigmoid ($k=6$) performed the best among all the other activation functions followed by Tangent-hyperbolic ($k=2$) function. Hence, no one activation function performed exceptionally well. Therefore, the efforts of selecting activation function, adaptively, by MOGP was essential in HFNTs performance. 

In this work, we were limited to examine the performance of our approach to only benchmark problems. Therefore, in presences of \textit{no free lunch theorem}~\cite{wolpert1997no,koppen2001remarks} and the algorithm's dependencies on random number generator, which are platforms, programming language, and implementation sensitive~\cite{l2005fast}, it is clear that performance of the mentioned approach is subjected to careful choice of training condition and parameter-setting when it comes to deal with other real-world problems.

\section{Conclusion}
\label{sec-con}
Effective use of the final population of the heterogeneous flexible neural trees (HFNTs) evolved using Pareto-based multiobjective genetic programming (MOGP) and the subsequent parameter tuning by differential evolution led to the formation of high-quality ensemble systems. The simultaneous optimization of accuracy, complexity, and diversity solved the problem of structural complexity that was inevitably imposed when a single objective was used. MOGP used in the \emph{tree construction} phase often guided an initial HFNT population towards a population in which the candidates were highly accurate, structurally simple, and diverse. Therefore, the selected candidates helped in the formation of a good ensemble system. The result obtained by HFNT$^\text{M}$ approach supports its superior performance over the algorithms collected for the comparison. In addition, HFNT$^\text{M}$ provides adaptation in structure, computational nodes, and input feature space. Hence, HFNT is an effective algorithm for automatic feature selection, data analysis, and modeling.          
	
\section*{Acknowledgment}
This work was supported by the IPROCOM Marie Curie Initial Training Network, funded through the People Programme (Marie Curie Actions) of the European Union’s Seventh Framework Programme FP7/2007–2013/, under REA grant agreement number 316555.

\section*{References}
\bibliography{fnt_mo_R2}	

\newpage
\appendix
\section{Dataset Description}
\begin{table}[!ht]
	\centering
	\footnotesize
	\caption{Collected datasets for testing HFNT$^\text{M}$}
	\label{tab-dataset}
	\setlength{\tabcolsep}{5pt}
	\begin{tabular}{clrrrl}
		\hline
		Index & Name & Features & Samples & Output & Type \\
		\hline
		AUS & Australia & 14 &  691 & 2 & \multirow{5}{*}{Classification}\\
		HRT & Heart     & 13 &  270 & 2 & \\
		ION & Ionshpere & 33 &  351 & 2 & \\
		PIM & Pima      &  8 &  768 & 2 & \\
		WDB & Wdbc      & 30 &  569 & 2 & \\
		\hline
		ABL & Abalone    &  8 &  4177 & 1 & \multirow{5}{*}{Regression}\\
		BAS & Baseball	 & 16 &   337 & 1 & \\
		DEE & DEE		 &  6 &   365 & 1 & \\			
		EVL & Elevators  & 18 & 16599 & 1 & \\
		FRD & Fridman    &  5 &  1200 & 1 & \\
		\hline   
		MGS & Mackey-Glass  &  4 & 1000  & 1 & \multirow{2}{*}{Time-series}\\
		WWR & Waste Water   & 4 & 475  & 1 & \\				   		
		\hline
	\end{tabular}
\end{table}

\section{Algorithms from literature}
\begin{table}[!ht]
	\centering
	\footnotesize 
	\caption{Algorithms from literature for the comparative study with HFNT$^\text{M}$}
	\label{tab_litModels}
	\setlength{\tabcolsep}{3pt}
	\begin{tabular}{r l l}
		\hline			 
		Ref. & Algorithms & Definition\\
		\hline
		\cite{haykin2009neural}         & MLP & Multi-layer Perceptron \\
		\cite{zhou2002hybrid}           & HDT & Hybrid Decision Tree \\ 	
		\cite{chen2006ensemble}         & FNT & Flexible Neural Tree  \\				   	
		\cite{jang1993anfis}		    & ANFIS-SUB & Adaptive Neuro-Fuzzy Inference System Using Subtractive Clustering \\
		\cite{cordon1999two}		    & TSK-IRL & Genetic Learning of TSK-rules Under Iterative Rule Learning \\	
		\cite{rustagi1994optimization}  & LINEAR-LMS & Least Mean Squares Linear Regression\\ 
		\cite{alcala2007local} 		    & LEL-TSK & Local Evolutionary Learning of TSK-rules\\  	
		\cite{cho1996radial} 			& RBF  & Classical Radial Basis Function\\
		\cite{van2004cooperative}		& CPSO & Cooperative Particle Swarm Optimization (PSO)\\
		\cite{psoBBNN}					& PSO-BBFN & PSO-based Beta Basis Function Neural Network\\
		\cite{gBBNN}  					& G-BBFNN  & GA-based BBFNN  \\
		\cite{juang2010hierarchical}    & HCMSPSO & Hierarchical Cluster-Based Multispecies PSO \\
		\cite{fwnn} 					& FWNN-M & Fuzzy Wavelet Neural Network Models\\
		\cite{dhahri2012hierarchical}	& HMDDE-BBFNN & Hierarchical Multidimensional DE-Based BBFNN\\
		\cite{miranian2013developing}	& LNF & Local Least-Squares Support Vector Machines-Based Neuro-Fuzzy Mode\\
		\cite{kasabov1996foundations}   & BPNN & Back-propagation Neural Network \\
		\cite{kasabov1999evolving}	    & EFuNNs & Evolving Fuzzy Neural Networks \\ 
		\cite{bouaziz2013extended}      & FBBFNT-EGP\&PSO & Extended Immune Programming and Opposite-PSO for Flexible BBFNN\\ 
		\cite{gacto2014metsk}		    & METSK-HD$^e$ & Multiobjective Evolutionary Learning of TSK-rules for High-Dimensional Problems \\           						   
		\hline
	\end{tabular}
\end{table}
\end{document}